%% file: main.tex
\documentclass[sigconf,nonacm]{acmart}
\ifdefined\pdfsuppresswarningpagegroup\pdfsuppresswarningpagegroup=1\fi

\AtBeginDocument{%
  }

\usepackage{booktabs}
\usepackage{array}
\usepackage{multirow}
\usepackage{xcolor}
\usepackage{subcaption}
\usepackage{tikz}
\usetikzlibrary{positioning,arrows.meta,fit,backgrounds,calc,shapes.misc}
\usepackage{fontawesome5}
\usepackage{enumitem}
\usepackage[noend]{algpseudocode}
\usepackage{algorithm}

\newcommand{\qera}{QERA}
\definecolor{cin}{HTML}{6C757D}
\definecolor{cask}{HTML}{3B6BA5}
\definecolor{cans}{HTML}{2A9D8F}
\definecolor{cchk}{HTML}{D98A0B}
\definecolor{crep}{HTML}{C8475B}
\definecolor{cvan}{HTML}{B5414F}

\begin{document}

\title{Ask Before You Answer: A Question-Driven Evidence Reasoning
  Architecture (QERA) for In-Context Learning}

\author{Jinhu Qi}
\affiliation{%
  \institution{The Chinese University of Hong Kong}
  \city{Hong Kong}
  \country{China}}
\email{jhqi25@cse.cuhk.edu.hk}

\author{Minda Hu}
\affiliation{%
  \institution{The Chinese University of Hong Kong}
  \city{Hong Kong}
  \country{China}}
\email{mdhu22@cse.cuhk.edu.hk}

\author{Wentao Zhang}
\affiliation{%
  \institution{Macao Polytechnic University}
  \city{Macao}
  \country{China}}
\email{p2522808@mpu.edu.mo}

\author{Zixing Song}
\affiliation{%
  \institution{The Hong Kong Polytechnic University}
  \city{Hong Kong}
  \country{China}}
\email{zixing.song@polyu.edu.hk}

\author{Irwin King}
\affiliation{%
  \institution{The Chinese University of Hong Kong}
  \city{Hong Kong}
  \country{China}}
\email{king@cse.cuhk.edu.hk}

\renewcommand{\shortauthors}{Qi et al.}

\input{sections/abstract}




\maketitle

\input{sections/intro}
\input{sections/related}
\input{sections/method}
\input{sections/experiments}
\input{sections/conclusion}

\nocite{dou2026clbenchlife}
\begin{acks}
We used large language model assistants for writing support, for preparing figures and tables, and for analysis code; all content was reviewed by the authors, who take full responsibility for it. The icons in Figures~1 and~2 are from Microsoft's Fluent Emoji (MIT License).
\end{acks}

\bibliographystyle{ACM-Reference-Format}
\bibliography{refs}

\appendix
\footnotesize
\input{sections/appendix}

\end{document}

%% file: sections/abstract.tex
\begin{abstract}
AI assistants are increasingly asked to reason over a user's own web traces: group chats, community threads, activity logs and revision histories. On CL-bench Life, a benchmark of such real-life context learning, frontier LLMs solve fewer than one task in five. We find that the request is not the specification: over 80\% of the benchmark's rubrics share less than 60\% of their content words with the final request, as for a rule set two turns earlier or every record that meets the user's criterion, and LLMs answering directly pass only 67\% of these implicit rubrics against 81\% of the others. We propose \qera{}, a Question-driven Evidence Reasoning Architecture that asks before it answers. \qera{} first turns the request into a requirement checklist and typed information needs, filtered by code for faithfulness to the user's words; it then answers from the whole conversation, asks the same questions again with quotes verified by code, and applies a guarded local repair. It needs no training or rubrics and makes about four LLM calls per task. On all 405 tasks with four frontier LLMs, \qera{} attains the highest solving rate on every LLM (21.0\% on average, against 18.2\% for the strongest of three inference-time baselines and 14.6\% for direct answering) and leads every baseline significantly when pooled across LLMs, while using fewer tokens than PSCI and 4--5\% of the tokens of Ctx2Skill. Ablations on GPT-6-Luna show that the gain is settled before the answer, by a fixed answer protocol (+3.8 rubric points) and the request-specific questions (+1.6), and that it is concentrated on implicit rubrics, with checking afterwards refining mainly numeric requirements.
\end{abstract}

%% file: sections/intro.tex
\section{Introduction}
\label{sec:intro}

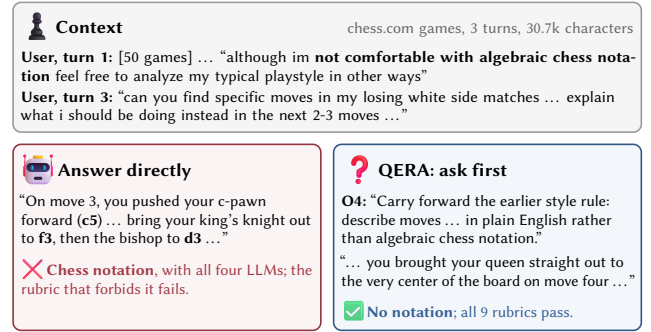
\begin{figure}[t]
  \centering
  \input{figures/fig_teaser}
  \caption{A CL-bench Life task (Game Logs). The user's first turn rules out chess notation; the request two turns later does not repeat it. Answering directly, all four LLMs write chess notation. \qera{}'s checklist carries the rule forward, and \qera{} solves the task with Kimi K3 and Grok 4.7 (both replies shown are from Kimi K3).}
  \Description{Top strip: answering directly sends the conversation through one LLM call to a reply that misses unstated requirements, with 81 percent pass rate on explicit and 67 percent on implicit rubrics. Bottom panel: a left-to-right pipeline from the conversation x through four numbered LLM stages, Ask, Answer, Check and Repair, with the artifacts Q, the draft, the audited evidence and the reply between them; dashed code boxes below each stage show the filters, the answer protocol, the measure-and-audit step and the guard; a blue line carries Q to Check and Repair.}
  \label{fig:teaser}
\end{figure}

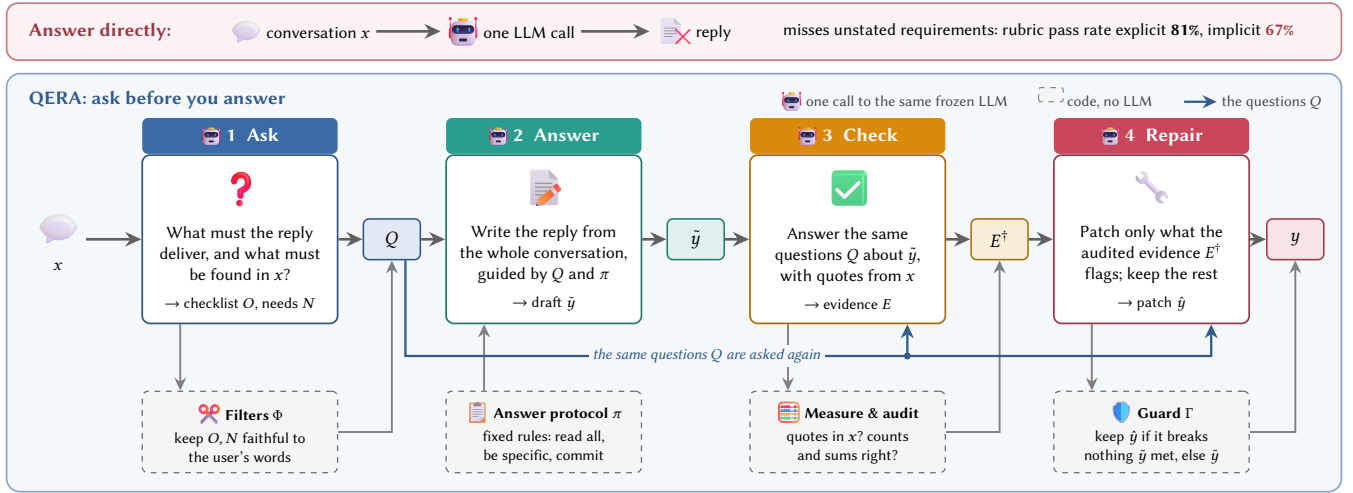
\begin{figure*}[t]
  \centering
  \resizebox{\textwidth}{!}{\input{figures/fig_arch}}
  \caption{Top: answering directly misses what a request leaves unsaid; the four LLMs of our study, answering directly, pass 81\% of the rubrics whose wording the request largely contains but only 67\% of the others. Bottom: \qera{} runs Eq.~(\ref{eq:qera}) from left to right. \textsc{Ask} writes the checklist $O$ and the information needs $N$, which the filters $\Phi$ turn into the questions $Q$; \textsc{Answer} drafts $\tilde y$ under $Q$ and the protocol $\pi$; \textsc{Check} answers $Q$ again with quotes, audited by code into $E^{\dagger}$; \textsc{Repair} patches only what $E^{\dagger}$ flags, and the guard $\Gamma$ keeps the patch $\hat y$ or the draft. The blue line carries the same $Q$ to \textsc{Check} and \textsc{Repair}.}
  \Description{Two panels. Left: answering directly, a conversation goes through one LLM call to a reply that misses unstated requirements, with bars showing 81 percent pass rate on explicit and 67 percent on implicit rubrics. Right: QERA's four stages Ask, Answer, Check and Repair, each with an icon and its question; below them the checklist and needs, the code filters producing the questions Q, the draft, the evidence, the measure-and-audit step, the guard and the final reply; arrows show Q feeding the draft and, dashed, the check.}
  \label{fig:arch}
\end{figure*}

People increasingly hand AI assistants the traces of their life on the Web: a group chat to summarize, two Reddit threads to compare, months of fitness logs or fifty chess.com games to analyze, the revision history of a shared document to reconcile. Answering such a request is in-context learning in its broad sense~\cite{brown2020gpt3,dou2026clbench}: the knowledge the reply needs lives in the context supplied at inference time, which here is messy, fragmented, spread over several people and turns, and personal. CL-bench Life~\cite{dou2026clbenchlife} collects 405 such tasks with 5,348 human-written rubrics; when it was released, the best of ten frontier LLMs solved 19.3\% of them, and the average was 13.8\%.

We find that much of the difficulty lies in the request rather than in the context. For over 80\% of CL-bench Life's rubrics, the final request contains less than 60\% of the rubric's content words: such rubrics test a rule the user set two turns earlier, a later message that corrects a fact, every record that meets the user's criterion, the person behind each claim. In one task, a user pastes fifty chess games and remarks that they are not comfortable with algebraic chess notation; asked two turns later for the moves they should have played, all four LLMs we test answer in chess notation (Figure~\ref{fig:teaser}). Answering directly, these LLMs pass 81\% of the remaining, explicit rubrics but only 67\% of these implicit ones. Zhong et al.~\cite{zhong2026psci} report a similar pattern on CL-bench: most of the specifications its rubrics check are unstated in the query yet traceable to the context. What the request leaves out is in the context, and it has to be recovered before the reply is written.

Existing inference-time methods for context learning recover it in other ways. CCA~\cite{qi2026cca} compiles the context into typed rules with executable verifiers, and Ctx2Skill~\cite{si2026ctx2skill} distills a skill for each context by multi-agent self-play; both organize the knowledge of the context. PSCI~\cite{zhong2026psci} comes closest to asking: before answering, it induces a specification contract for the task, with the help of rubric-style demonstrations from other tasks, and then checks and repairs the answer against it. None of them turns the request into typed questions about the context (which records to enumerate, which facts to look up, which figures to compute) and reuses those questions to check the answer against quoted evidence. On CL-bench Life, the strongest of them changes from one LLM to the next, and Ctx2Skill spends 2.4M--3.1M tokens per task, because a context that serves a single request cannot amortize its self-play.

Reference librarians have long held that a patron's stated question is a compromised version of the actual information need, and that a good answer begins by negotiating the question~\cite{taylor1968question}; search systems ask clarifying questions for the same reason~\cite{aliannejadi2019asking,zamani2020generating}. An assistant that reads a user's traces cannot interview the user about every request, but it can interview the context. We propose \textbf{\qera{}}, a \textbf{Q}uestion-driven \textbf{E}vidence \textbf{R}easoning \textbf{A}rchitecture that asks before it answers (Figure~\ref{fig:arch}). Its \emph{Ask} stage turns the request into a requirement checklist (what must the reply deliver, including rules from earlier turns?) and typed information needs (what must be looked up, enumerated, counted, computed, compared or judged in the context?), which code filters keep faithful to the user's words. The \emph{Answer} stage writes the reply from the whole conversation, guided by these questions and a fixed answer protocol; the \emph{Check} stage asks the same questions again, answered with quotes that code verifies; and a \emph{Guarded Repair} patches only what fails. \qera{} needs no training or rubrics and makes about four LLM calls per task.

We evaluate \qera{} on all 405 tasks of CL-bench Life with four frontier LLMs and four baselines, around three research questions:
\begin{itemize}[leftmargin=*,topsep=2pt,itemsep=1pt]
  \item \textbf{RQ1 (Effectiveness and cost):} Does asking before answering improve real-life context learning across LLMs, and at what cost relative to strong inference-time methods?
  \item \textbf{RQ2 (Asking vs.\ checking):} Which stages produce the gain: the questions asked before the answer, or the evidence check after it?
  \item \textbf{RQ3 (Source of the gains):} Which requirements and contexts benefit? Are the gains concentrated on implicit requirements, and do they hold across context lengths and types?
\end{itemize}

\noindent Our contributions answer them in turn:
\begin{itemize}[leftmargin=*,topsep=2pt,itemsep=1pt]
  \item \textbf{(RQ1)} We propose \qera{}, a question-driven inference-time architecture for learning from real-life context. It attains the highest solving rate on all four LLMs (21.0\% on average, against 18.2\% for the strongest baseline and 14.6\% for direct answering), leads every baseline significantly when pooled across LLMs, and uses fewer tokens than PSCI and 4--5\% of the tokens of Ctx2Skill.
  \item \textbf{(RQ2)} A stage-wise ablation shows that the gain is settled before the answer: on GPT-6-Luna, the fixed answer protocol adds 3.8 rubric points and the request-specific questions 1.6 (89\% of the gain), and \qera{}'s two-call draft already solves more tasks than every baseline on every LLM; checking afterwards refines mainly numeric requirements.
  \item \textbf{(RQ3)} The gains are concentrated on implicit rubrics (2.5 points above the best baseline and 6.6 above direct answering, while all methods lie within 1.4 points of each other on the rest), and \qera{} ranks first in every context-length quartile and context category and in 20 of the 36 combinations of LLM and subcategory.
\end{itemize}

%% file: figures/fig_teaser.tex
\providecommand{\ico}[2][0.9cm]{\includegraphics[height=#1]{figures/icons/#2}}
\begin{tikzpicture}[x=1cm, y=1cm, font=\sffamily\scriptsize,
  box/.style={rounded corners=3pt, line width=0.6pt, inner sep=3pt, anchor=north west, align=left},
]
\node[box, draw=black!40, fill=black!4, text width=8.1cm] (c) at (0,0) {%
  \raisebox{-0.3\height}{\ico[0.42cm]{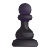}}\,{\footnotesize\bfseries Context}\hfill\textcolor{black!60}{chess.com games, 3 turns, 30.7k characters}\\[2pt]
  \textbf{User, turn 1:} [50 games] \ldots\ ``although im \textbf{not comfortable with algebraic chess notation} feel free to analyze my typical playstyle in other ways''\\[1pt]
  \textbf{User, turn 3:} ``can you find specific moves in my losing white side matches \ldots\ explain what i should be doing instead in the next 2-3 moves \ldots''};
\node[box, draw=cvan!80!black, fill=cvan!6] (v) at ([yshift=-4pt]c.south west) {\begin{minipage}[t][2.25cm][t]{3.86cm}\raggedright
  \raisebox{-0.3\height}{\ico[0.45cm]{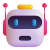}}\,{\footnotesize\bfseries Answer directly}\\[2pt]
  ``On move 3, you pushed your c-pawn forward (\textbf{c5}) \ldots\ bring your king's knight out to \textbf{f3}, then the bishop to \textbf{d3} \ldots''\\[3pt]
  \raisebox{-0.25\height}{\ico[0.3cm]{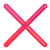}}\,\textcolor{cvan!90!black}{\textbf{Chess notation}, with all four LLMs; the rubric that forbids it fails.}\end{minipage}};
\node[box, draw=cask!80!black, fill=cask!6, anchor=north east] (q) at ([yshift=-4pt]c.south east) {\begin{minipage}[t][2.25cm][t]{3.86cm}\raggedright
  \raisebox{-0.3\height}{\ico[0.45cm]{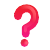}}\,{\footnotesize\bfseries \qera{}: ask first}\\[2pt]
  \textbf{O4:} ``Carry forward the earlier style rule: describe moves \ldots\ in plain English rather than algebraic chess notation.''\\[2pt]
  ``\ldots\ you brought your queen straight out to the very center of the board on move four \ldots''\\[3pt]
  \raisebox{-0.25\height}{\ico[0.3cm]{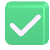}}\,\textcolor{cask!90!black}{\textbf{No notation}; all 9 rubrics pass.}\end{minipage}};
\end{tikzpicture}

%% file: figures/fig_arch.tex
\providecommand{\ico}[2][0.9cm]{\includegraphics[height=#1]{figures/icons/#2}}
\begin{tikzpicture}[x=1cm, y=1cm, font=\sffamily\normalsize,
  stage/.style={draw=#1!78!black, fill=white, rounded corners=4pt, line width=0.9pt, inner sep=4pt,
                minimum width=3.6cm, minimum height=3.1cm, align=center, anchor=center},
  head/.style={fill=#1, text=white, rounded corners=3pt, font=\sffamily\large\bfseries, inner sep=4pt,
               minimum width=3.6cm, anchor=south},
  art/.style={draw=#1!78!black, fill=#1!15, rounded corners=3pt, line width=0.8pt, inner sep=3pt,
              minimum width=1.05cm, minimum height=0.75cm, align=center, anchor=center, font=\sffamily\large\bfseries},
  code/.style={draw=black!55, dashed, fill=black!4, rounded corners=3pt, line width=0.7pt, inner sep=3pt,
               minimum width=3.6cm, minimum height=1.5cm, align=center, anchor=center, font=\sffamily\small},
  flow/.style={-{Stealth[length=3mm,width=2.7mm]}, line width=1.4pt, draw=black!62},
  aux/.style={-{Stealth[length=2.4mm,width=2.2mm]}, line width=1pt, draw=black!50},
  qline/.style={-{Stealth[length=2.6mm,width=2.4mm]}, line width=1.2pt, draw=cask!85!black},
  note/.style={font=\sffamily\small, text=black!78, align=center},
]
\def\xA{4.3} \def\xB{9.9} \def\xC{15.5} \def\xD{21.1}
\def\yS{-4.35}   %
\def\yQ{-6.5}    %
\def\yK{-7.9}    %
\fill[cvan!7, rounded corners=6pt] (0,0) rectangle (24.6,-1.05);
\draw[cvan!70, line width=0.9pt, rounded corners=6pt] (0,0) rectangle (24.6,-1.05);
\node[font=\sffamily\large\bfseries, text=cvan!85!black, anchor=west] at (0.3,-0.52) {Answer directly:};
\node[anchor=west] (v1) at (4.0,-0.52) {\raisebox{-0.3\height}{\ico[0.6cm]{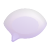}}\ conversation $x$};
\node[anchor=west] (v2) at (8.0,-0.52) {\raisebox{-0.3\height}{\ico[0.6cm]{robot}}\ one LLM call};
\node[anchor=west] (v3) at (11.85,-0.52) {\raisebox{-0.3\height}{\ico[0.55cm]{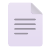}}\hspace{-6pt}\raisebox{-0.45\height}{\ico[0.32cm]{cross_mark}}\ reply};
\draw[flow] (v1.east) -- (v2.west);
\draw[flow] (v2.east) -- (v3.west);
\node[anchor=west, font=\sffamily\normalsize] at (14.2,-0.52) {misses unstated requirements: rubric pass rate explicit \textbf{81\%}, implicit \textbf{\textcolor{cvan!90!black}{67\%}}};
\fill[cask!5, rounded corners=7pt] (0,-1.3) rectangle (24.6,-9.0);
\draw[cask!60, line width=0.9pt, rounded corners=7pt] (0,-1.3) rectangle (24.6,-9.0);
\node[font=\sffamily\large\bfseries, text=cask!85!black, anchor=west] at (0.3,-1.78) {\qera{}: ask before you answer};
\node[font=\sffamily\small, text=black!72, anchor=east] at (24.35,-1.78)
  {\raisebox{-0.3\height}{\ico[0.4cm]{robot}}\ one call to the same frozen LLM\qquad
   \tikz[baseline=-0.6ex]\draw[black!55, dashed, fill=black!4, rounded corners=1pt] (0,0) rectangle (0.45,0.26);\ code, no LLM\qquad
   \tikz[baseline=-0.6ex]\draw[qline] (0,0) -- (0.6,0);\ the questions $Q$};
\node (xin) at (0.95,\yS+0.2) {\ico[0.8cm]{speech_balloon}};
\node at (0.95,\yS-0.5) {$x$};
\node[stage=cask] (ask) at (\xA,\yS) {\ico[0.75cm]{red_question_mark}\\[4pt]
  What must the reply\\deliver, and what must\\be found in $x$?\\[4pt]{\small$\to$ checklist $O$, needs $N$}};
\node[head=cask] at (ask.north) {\raisebox{-0.25\height}{\ico[0.38cm]{robot}}\ 1\;\;Ask};
\node[stage=cans] (ans) at (\xB,\yS) {\ico[0.75cm]{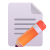}\\[4pt]
  Write the reply from\\the whole conversation,\\guided by $Q$ and $\pi$\\[4pt]{\small$\to$ draft $\tilde y$}};
\node[head=cans] at (ans.north) {\raisebox{-0.25\height}{\ico[0.38cm]{robot}}\ 2\;\;Answer};
\node[stage=cchk] (chk) at (\xC,\yS) {\ico[0.75cm]{check_mark_button}\\[4pt]
  Answer the same\\questions $Q$ about $\tilde y$,\\with quotes from $x$\\[4pt]{\small$\to$ evidence $E$}};
\node[head=cchk] at (chk.north) {\raisebox{-0.25\height}{\ico[0.38cm]{robot}}\ 3\;\;Check};
\node[stage=crep] (rep) at (\xD,\yS) {\ico[0.75cm]{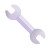}\\[4pt]
  Patch only what the\\audited evidence $E^{\dagger}$\\flags; keep the rest\\[4pt]{\small$\to$ patch $\hat y$}};
\node[head=crep] at (rep.north) {\raisebox{-0.25\height}{\ico[0.38cm]{robot}}\ 4\;\;Repair};
\node[art=cask] (Q)  at ({(\xA+\xB)/2},\yS) {$Q$};
\node[art=cans] (dr) at ({(\xB+\xC)/2},\yS) {$\tilde y$};
\node[art=cchk] (E)  at ({(\xC+\xD)/2},\yS) {$E^{\dagger}$};
\node[art=crep] (y)  at (23.75,\yS) {$y$};
\draw[flow] (xin.east |- 0,\yS) -- (ask.west);
\draw[flow] (ask.east) -- (Q.west);
\draw[flow] (Q.east) -- (ans.west);
\draw[flow] (ans.east) -- (dr.west);
\draw[flow] (dr.east) -- (chk.west);
\draw[flow] (chk.east) -- (E.west);
\draw[flow] (E.east) -- (rep.west);
\draw[flow] (rep.east) -- (y.west);
\node[code] (phi)   at (\xA,\yK) {\raisebox{-0.25\height}{\ico[0.45cm]{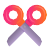}}\ \textbf{Filters $\Phi$}\\[2pt]keep $O,N$ faithful to\\the user's words};
\node[code] (proto) at (\xB,\yK) {\raisebox{-0.25\height}{\ico[0.45cm]{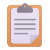}}\ \textbf{Answer protocol $\pi$}\\[2pt]fixed rules: read all,\\be specific, commit};
\node[code] (aud)   at (\xC,\yK) {\raisebox{-0.25\height}{\ico[0.45cm]{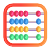}}\ \textbf{Measure \& audit}\\[2pt]quotes in $x$? counts\\and sums right?};
\node[code] (gd)    at (\xD,\yK) {\raisebox{-0.25\height}{\ico[0.45cm]{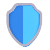}}\ \textbf{Guard $\Gamma$}\\[2pt]keep $\hat y$ if it breaks\\nothing $\tilde y$ met, else $\tilde y$};
\draw[aux] ([xshift=-1.1cm]ask.south) -- ([xshift=-1.1cm]phi.north);
\draw[aux] ([xshift=-1.1cm]proto.north) -- ([xshift=-1.1cm]ans.south);
\draw[aux] ([xshift=-1.1cm]chk.south) -- ([xshift=-1.1cm]aud.north);
\draw[aux] ([xshift=-1.1cm]rep.south) -- ([xshift=-1.1cm]gd.north);
\draw[aux] (phi.east) -| (Q.south);
\draw[aux] (aud.east) -| (E.south);
\draw[aux] (gd.east) -| (y.south);
\draw[qline] ([xshift=0.25cm]Q.south) |- (\xC+1.1,\yQ) -- ([xshift=1.1cm]chk.south);
\draw[qline] (\xC+1.1,\yQ) -- (\xD+1.1,\yQ) -- ([xshift=1.1cm]rep.south);
\fill[cask!85!black] (\xC+1.1,\yQ) circle (1.8pt);
\node[font=\sffamily\small\itshape, text=cask!85!black, fill=cask!5, inner sep=2pt] at ({(\xB+\xC)/2+0.2},\yQ) {the same questions $Q$ are asked again};
\end{tikzpicture}

%% file: sections/related.tex
\section{Related Work}
\label{sec:related}

\paragraph{Learning from context}
In-context learning originally denoted learning a task from demonstrations in the prompt~\cite{brown2020gpt3}. Context learning benchmarks widen it to learning new knowledge, rules and procedures from long, task-specific contexts at inference time~\cite{dou2026clbench}, and CL-bench Life moves it to real-life context such as chats, personal archives and behavioral traces~\cite{dou2026clbenchlife}. Long-context studies show that models use long inputs unevenly~\cite{liu2024lost,hsieh2024ruler,bai2024longbench}, multi-turn studies show that performance drops sharply when a task's requirements are revealed over several turns~\cite{laban2026lost}, and instruction-following benchmarks test explicit, verifiable constraints~\cite{zhou2023ifeval}. Work on conversational memory and personalization builds and evaluates assistants that store or retrieve a user's history~\cite{packer2023memgpt,maharana2024locomo,wu2025longmemeval,salemi2024lamp,li2024personal}. \qera{} neither retrieves nor stores: the whole conversation stays in every call, and the difficulty it addresses is recognizing what a request leaves implicit.

\paragraph{Inference-time methods for context learning}
Context engineering adapts what a model sees without changing its weights~\cite{mei2025context}. ACE~\cite{zhang2026ace} and Dynamic Cheatsheet~\cite{suzgun2026dynamic} accumulate strategies across tasks; Ctx2Skill~\cite{si2026ctx2skill} discovers skills for each context by self-play; CCA~\cite{qi2026cca} compiles the context into typed rules with executable verifiers; PSCI~\cite{zhong2026psci} induces a specification contract, then checks and repairs the answer. CCA and Ctx2Skill organize the knowledge of each context. PSCI is closest to our work: it induces, before answering, a contract of the specifications a correct answer must meet, and then checks and repairs the answer, a shape that \qera{}'s later stages share. \qera{}'s questions differ: they are anchored to the user's words and filtered by code, need no demonstrations or other tasks, and also ask what must be found in the context, as typed information needs whose findings are verified against the conversation and reused to check the reply.

\paragraph{Asking, decomposing and verifying}
Reasoning step by step~\cite{wei2022cot} or through sub-questions and sub-tasks solved on the way to the answer improves LLM reasoning~\cite{press2023selfask,zhou2023leasttomost,wang2023planandsolve,khot2023decomposed}, and ReAct interleaves reasoning with actions~\cite{yao2023react}. In information retrieval, systems ask users clarifying questions to recover underspecified needs~\cite{aliannejadi2019asking,zamani2020generating}, and query expansion rewrites a query with generated text~\cite{wang2023query2doc,gao2023hyde}. \qera{}'s questions are addressed to the context rather than to the user or a retriever, they are typed so that enumerations and computations can be audited, and they are asked twice, before and after the answer. On the verification side, self-refinement and self-reflection~\cite{madaan2023selfrefine,shinn2023reflexion}, chain-of-verification~\cite{dhuliawala2024cove} and tool-interactive critiquing~\cite{gou2024critic} revise answers after they are written, and program-aided reasoning delegates arithmetic to code~\cite{gao2023pal,chen2023pot}. LLMs, however, struggle to correct their own reasoning without external feedback~\cite{huang2024cannot}, and our RQ2 echoes this: once the right questions have been asked before the answer, most of the work is done.

%% file: sections/method.tex
\section{QERA: Ask Before You Answer}
\label{sec:method}

\subsection{Problem Formulation}
\label{sec:formulation}

A task is a conversation $x=(m_1,\dots,m_T)$ that ends with a user turn $m_T$. That turn holds the request $q$, often after pasted materials; the context $C$ is everything else the reply draws on: pasted chats, threads, logs, records and documents, and, in the 59.8\% of CL-bench Life tasks that are multi-turn, earlier rounds between the user and an assistant~\cite{dou2026clbenchlife}. A solver maps $x$ to a reply $y$. Each task $t$ comes with rubrics $r_{t,1},\dots,r_{t,n_t}$ written by annotators (13.2 per task on average), and a judge $J$ decides whether a reply satisfies each of them, $J(y,r)\in\{0,1\}$. Over a task set $T$ with replies $y_t$, we report the benchmark's metric, the \emph{solving rate}, and, as a secondary metric, the \emph{rubric pass rate}, the share of all rubrics that the replies satisfy:
\begin{equation}
  \mathrm{Solv.}=\frac{1}{|T|}\sum_{t\in T}\prod_{i=1}^{n_t} J(y_t,r_{t,i}), \qquad \mathrm{Rub.}=\frac{\sum_{t\in T}\sum_{i=1}^{n_t} J(y_t,r_{t,i})}{\sum_{t\in T} n_t}.
  \label{eq:metrics}
\end{equation}
The solver never sees the rubrics.

\paragraph{The request is not the specification}
We call a rubric \emph{explicit} when at least 60\% of its content words (names, numbers, quoted strings and words of five or more letters) occur in the request $q$, and \emph{implicit} otherwise (Appendix~\ref{app:full}). By this rule, 4,352 of the 5,333 rubrics it can classify (82\%) are implicit. The rule is a lexical proxy, and our conclusions hold at every level of overlap (\S\ref{sec:rq3}). Implicit rubrics test, for example, a rule the user set in an earlier turn, a later message that corrects a fact, every record that meets the user's criterion, the person behind each claim, the figure that the user's own numbers imply. They are also where frontier LLMs fail most: answering directly, the four LLMs of our study pass 81.1\% of the explicit rubrics but only 67.4\% of the implicit ones (\S\ref{sec:rq3}). The part of the specification that the request omits is in $C$, and a solver has to recover it before it can answer well.

\paragraph{Approach}
\qera{} recovers this specification as explicit questions before answering and asks the same questions again after answering. One pass over a task is a single pipeline (Figure~\ref{fig:arch}, Algorithm~\ref{alg:qera}):
\begin{equation}
\begin{aligned}
x \;\xrightarrow{\;\textsc{Ask}\;}\; (O,N) \;\xrightarrow{\;\Phi\;}\; Q
 \;&\xrightarrow{\;\textsc{Answer}_Q\;}\; \tilde y\\
 &\xrightarrow{\;\textsc{Check}_Q\;}\; E \;\xrightarrow{\;\textsc{Repair}_Q\;}\; \hat y \;\xrightarrow{\;\Gamma\;}\; y .
\end{aligned}
\label{eq:qera}
\end{equation}
Each stage in small capitals is one call to the same frozen LLM $\mathcal M$ with the whole conversation $x$ in its prompt (\textsc{Ask}, \textsc{Answer} and \textsc{Check} are repeated when an output is empty, cut off or malformed); Greek letters are code. \textsc{Ask} returns a requirement checklist $O$ and information needs $N$, and the filters $\Phi$ turn them into the questions $Q$. The subscript marks the stages that use $Q$: the draft $\tilde y$ is written under it, the check asks it again and returns evidence $E$, which code audits into $E^{\dagger}$, and the repair fixes only what $E^{\dagger}$ flags. The guard $\Gamma$ returns the patch $\hat y$ if it passes deterministic checks against $\tilde y$, and $\tilde y$ otherwise.

\begin{algorithm}[t]
\caption{\qera{} on one task}
\label{alg:qera}
\small
\begin{algorithmic}[1]
\Require conversation $x$; frozen LLM $\mathcal M$; answer protocol $\pi$
\Ensure reply $y$
\State $(O,N)\gets\mathcal M(x,\textsf{ask})$ \Comment{\textsc{Ask}: one JSON call}
\State $Q\gets\Phi(O,N;x)$ \Comment{code: filters, coverage}
\State $\tilde y\gets\mathcal M(x,\pi,Q)$ \Comment{\textsc{Answer}}
\State $m\gets\textsc{Measure}(\tilde y,x)$ \Comment{code: counts, limits}
\State $E\gets\mathcal M(x,Q,m,\tilde y)$ \Comment{\textsc{Check}: one JSON call}
\State $E^{\dagger}\gets\textsc{Audit}(E,Q,m,\tilde y,x)$ \Comment{code: quotes, sums}
\If{$E^{\dagger}=\emptyset$} \State \Return $\tilde y$ \EndIf
\State $\hat y\gets\mathcal M(x,O,E^{\dagger},m,\tilde y)$ \Comment{\textsc{Repair}}
\State \Return $\hat y$ \textbf{if} $\Gamma(\tilde y,\hat y)$ \textbf{else} $\tilde y$ \Comment{code: guard}
\end{algorithmic}
\end{algorithm}

\begin{table}[t]
\caption{The questions $Q$: what each asks for, what \textsc{Check} returns for it, and what code then verifies. Findings whose quote is not in $x$ are flagged for the repair and ignored by the audit. Shares are of all needs over the four LLMs.}
\label{tab:questions}
\centering
\footnotesize
\setlength{\tabcolsep}{2.4pt}
\renewcommand{\arraystretch}{0.95}
\begin{tabular}{@{}>{\raggedright\arraybackslash}p{1.35cm}>{\raggedright\arraybackslash}p{2.05cm}>{\raggedright\arraybackslash}p{2.1cm}>{\raggedright\arraybackslash}p{2.3cm}@{}}
\toprule
Question & Asks for & \textsc{Check} returns & Code verifies \\
\midrule
requirement $o_k$ & what the reply must deliver & verdict and a quote of the draft & measured lengths and counts; required strings; exclusions \\
\midrule
lookup (28\%) & a specific fact & fact, quote, location & quote in $x$; found values in $\tilde y$ \\
enumerate (47\%) & every item meeting the user's criterion & one finding per item & items of a need called correct that $\tilde y$ never names \\
count (3\%) & a number over items & the items and the count & count $=$ number of items \\
compute (4\%) & arithmetic on stated values & every input and the result & inputs sum or average to the result \\
compare (5\%) & a comparison & facts on each side & quotes in $x$; found values in $\tilde y$ \\
judge (13\%) & an inference from evidence & evidence and verdict & quotes in $x$; found values in $\tilde y$ \\
\bottomrule
\end{tabular}
\end{table}

\subsection{Ask: Questions Before the Answer}
\label{sec:ask}

One structured call returns two sets of questions and answers none of them:
\begin{equation}
\begin{aligned}
O&=\{o_k\}_{k=1}^{K}, & o_k&=(u_k,\,s_k,\,\kappa_k),\\
N&=\{n_j\}_{j=1}^{J}, & n_j&=(d_j,\,\tau_j,\,R_j,\,w_j).
\end{aligned}
\label{eq:questions}
\end{equation}
A requirement $o_k$ is what the reply must deliver: $u_k$ are the user's words that create it, $s_k$ is its scope (people, items, period, units, and any criterion or definition the user gave, copied verbatim), and $\kappa_k$ its kind (list, count, compare, infer, write, \ldots). $O$ also records the answer format, required strings, exclusions grounded in the user's words or in the request's stated purpose, and the layout of an earlier reply that the request edits; rules set in earlier turns are carried forward. A need $n_j$ is a fact to find or derive in $C$: a description $d_j$ in the user's words, a type $\tau_j$ (Table~\ref{tab:questions}), the requirements $R_j\subseteq O$ it serves, and where to start looking, $w_j$. A list or count requirement first gets an enumerate need over all of the materials, and further needs ask for later corrections. \textsc{Ask} returns 4.2 requirements and 7.6 needs per task on average (7.8 needs after $\Phi$).

\paragraph{Faithfulness filters $\Phi$}
A question the user never implied makes the answer hedge, narrow or reformat. $\Phi$ therefore removes every part of $O$ and $N$ that the user's words $U(x)$ do not support: exclusions that quote words absent from $U(x)$, required strings absent from $U(x)$, caution and brevity clauses that no user turn asks for, date windows the user never set, unrequested formats, and stale templates (Appendix~\ref{app:filters}). It then adds missing coverage:
\begin{equation}
\begin{split}
N\leftarrow N\cup\big\{\nu(o):\;& o\in O,\ \kappa(o)\in\mathcal K_{\mathrm{set}},\\
&\nexists\, n\in N:\ o\in R(n),\ \tau(n)\in\mathcal T_{\mathrm{set}}\big\},
\end{split}
\label{eq:cover}
\end{equation}
where $\mathcal K_{\mathrm{set}}$ holds list, count-or-compute, timeline, summary and classification requirements, $\mathcal T_{\mathrm{set}}=\{\text{enumerate},\text{count},\text{compute}\}$, and $\nu(o)$ asks for every item that the user's words in $o$ cover (12--16\% of tasks).

\subsection{Answer: Write from the Whole Conversation}
\label{sec:answer}

The draft $\tilde y=\mathcal M(x,\pi,Q)$ is written in one call over the whole conversation, so no evidence is lost to retrieval or extraction. $Q$ enters as an advisory plan: the user's words and the materials override it, and anything it misses must still be covered. The answer protocol $\pi$ is fixed and settles the questions every request raises (Appendix~\ref{app:prompts}): (i)~read every part that bears on the request, including later corrections; (ii)~give the specifics the materials give, not category words; (iii)~include every qualifying item, with no threshold of one's own; (iv)~open yes/no, count and ranking answers with a committed verdict; (v)~show the inputs of every total, share or rate; (vi)~keep attributions, format, length, language and exclusions, without disclaimers.

\subsection{Check: Ask the Same Questions Again}
\label{sec:check}

Code first measures the draft ($m$: word, sentence and item counts, the user's limits, missing strings and headings, excluded items). One structured call then answers every question in $Q$ against $\tilde y$ and the whole conversation:
\begin{equation}
E=\big\{(o_k,\,v_k,\,a_k)\big\}_{k=1}^{K}\;\cup\;\big\{(n_j,\,F_j,\,\sigma_j)\big\}_{j=1}^{J},
\label{eq:evidence}
\end{equation}
with a verdict $v_k$ (pass, partial or fail) backed by a quote $a_k$ of the draft, findings $F_j$ (each a fact with a quote of under 25 words and its location; one per item for enumerate and count needs, one per input for compute needs), and a status $\sigma_j$ (correct, missing, incomplete, wrong, or not applicable). $E$ also lists latent gaps, unsupported facts, format violations and a plan of at most twelve edits. Three rules keep the check from doing harm: a count is wrong only if the check names the item missed or added; a requested inference that rests on stated evidence is not unsupported; and no edit may hedge a committed answer. Code then audits $E$: findings whose quote does not occur in $x$ are flagged so that the repair may not rely on them, and each typed need is recomputed as in the last column of Table~\ref{tab:questions}. $E^{\dagger}$ collects the check's actionable items, the audit's findings, and any missing required string, excluded item or leaked label in $m$; a check that returns nothing usable leaves $E^{\dagger}=\emptyset$.

\subsection{Guarded Repair}
\label{sec:repair}

If $E^{\dagger}\neq\emptyset$, one more call patches only what fails, $\hat y=\mathcal M(x,O,E^{\dagger},m,\tilde y)$, where the needs enter through their findings in $E^{\dagger}$: it applies each fix with the check's facts and quotes, skips fixes the materials contradict, keeps definite answers definite, propagates changed numbers, and preserves the rest. The guard then decides
\begin{equation}
y=\begin{cases}\hat y & \text{if } \bigwedge_{g\in\mathcal G} g(\tilde y,\hat y),\\ \tilde y & \text{otherwise,}\end{cases}
\label{eq:guard}
\end{equation}
where $\mathcal G$ holds deterministic checks against the draft: the patch is non-empty and not cut off; unless a length fix was requested, it keeps at least half of the draft's length and 30\% of its numbers; it adds no year outside the conversation; and it keeps every word limit, count, required string and exclusion that the draft met (Appendix~\ref{app:filters}). A code pass finally restores quotations to their source text. \qera{} thus makes three LLM calls when nothing is flagged and four otherwise (3.8--4.0 per task, retries included; Appendix~\ref{app:cost}), and uses no training, demonstrations or rubrics.

%% file: sections/experiments.tex
\section{Experiments}
\label{sec:experiments}

\begin{table*}[t]
  \caption{Results on all 405 CL-bench Life tasks (135 per category), graded by GPT-5.1: solving rate (Solv.) and rubric pass rate (Rub.) in \%, for the three context categories (Communication \& Social Interactions, Behavioral Records \& Activity Trails, Fragmented Information \& Revisions) and overall, and API tokens (input + output) per task. Within each block, \textbf{bold} is the best and \underline{underlined} the second best value. $^{*}$On GPT-6-Luna, Ctx2Skill reuses the skills selected by its Grok 4.7 self-play (\S\ref{sec:setup}); its tokens are that self-play's (3.10M Grok 4.7 tokens per task) plus GPT-6-Luna's inference (28k).}
  \label{tab:main}
  \centering
  \input{tables/main}
\end{table*}

\subsection{Experimental Setup}
\label{sec:setup}

\paragraph{Benchmark}
We use all 405 tasks of CL-bench Life~\cite{dou2026clbenchlife}, which pose requests over 404 distinct real-life contexts and come with 5,348 rubrics written by annotators. The tasks fall into three categories of 135 tasks and nine subcategories of 45: \emph{Communication \& Social Interactions} (private conversations, group conversations and meeting transcripts, community interactions), \emph{Behavioral Records \& Activity Trails} (digital footprints and daily-life records, game logs, self-tracking trajectories), and \emph{Fragmented Information \& Revisions} (creation and revision histories, personal and public information fragments). Contexts are 19.4K tokens long on average and up to 170.8K.

\paragraph{LLMs}
We use four frontier reasoning LLMs from four providers: Kimi~K3 from Moonshot AI~\cite{moonshot2026kimik3}, Grok~4.7 from xAI~\cite{xai2026grok47} and GLM-5 from Z.ai~\cite{zai2026glm5}, all served through Amazon Bedrock, and GPT-6-Luna from OpenAI (version 2026-09-22)~\cite{openai2026gpt6luna}, served through Azure OpenAI. Every call of every method uses high reasoning effort and the model's full output limit, so that truncation never separates methods, and all calls of a method in a run use the same frozen model.

\paragraph{Baselines}
\emph{Vanilla} sends the conversation to the model once. \emph{CCA}~\cite{qi2026cca} compiles the context into a typed representation of rules, output specification and data profile, generates executable verifiers, and corrects the answer when they flag violations; we run the authors' released pipeline. \emph{PSCI}~\cite{zhong2026psci}, which reports the best results on CL-bench (28.1\% solved with GPT-5.1), induces a specification contract from the context and from rotated rubric-style demonstrations, generates under it, checks the answer adversarially and repairs it; no code is released, so we re-implement it with the prompt templates of its paper, verbatim, and draw its demonstrations from CL-bench~\cite{dou2026clbench}, never from CL-bench Life. \emph{Ctx2Skill}~\cite{si2026ctx2skill} discovers a skill for each context by multi-agent self-play and injects the selected skill at inference; we run the official code with every role played by the evaluated LLM. Ctx2Skill costs 2.4M--3.1M tokens per task on CL-bench Life, almost all of them in self-play, so on GPT-6-Luna we reuse the skills that the Grok 4.7 self-play selected for each context and run only the final generation on GPT-6-Luna (Ctx2Skill$^{*}$); we charge it the full cost, the reused self-play plus the GPT-6-Luna inference. Appendix~\ref{app:baselines} gives every implementation detail.

\paragraph{Evaluation}
Every reply is graded by the benchmark's judge model, GPT-5.1 at high reasoning effort, which decides each rubric (Eq.~\ref{eq:metrics}) with the benchmark's grading template; we replace the template's opening role sentence, which the API rejected as a policy violation, with a neutral instruction and keep the rest, with its per-rubric instructions and output format, verbatim (Appendix~\ref{app:baselines}). The judge sees only the reply and the rubrics, never the conversation or the method that produced the reply. The benchmark's authors validated this judge: in a manual check of 100 sampled solutions, its verdicts were over 90\% accurate, and on the outputs of five models GPT-5.1 agreed with Claude Opus 4.6 and Gemini 3.1 Pro on 93.2\% and 93.0\% of rubric verdicts (Cohen's $\kappa$ 0.72 and 0.71)~\cite{dou2026clbenchlife}. API failures are retried for every method, and a task left without an answer counts as failed (4 of 8,100 runs; Appendix~\ref{app:baselines}). We compare methods with paired bootstrap tests over tasks~\cite{efron1994bootstrap} and report 95\% intervals in brackets; pooled comparisons resample tasks together with their four LLM results (405 tasks $\times$ 4 LLMs = 1,620 pairs). Differences are computed from unrounded rates. Tokens are the input and output tokens reported by each provider, reasoning tokens included.

\paragraph{No rubric access}
No method observes rubrics at inference, and \qera{}'s prompts contain no task-specific content: a single set of prompts serves every task and every LLM (Appendix~\ref{app:prompts}).

\subsection{RQ1: Effectiveness and Cost}
\label{sec:rq1}

\begin{table}[t]
  \caption{\qera{} minus each baseline, pooled over 405 tasks $\times$ 4 LLMs, in percentage points of unrounded rates, with paired 95\% bootstrap intervals (tasks resampled with their four LLM results). Ctx2Skill includes Ctx2Skill$^{*}$ on GPT-6-Luna. All eight differences remain significant after Holm correction (adjusted $p\le0.014$). Per-LLM tests: Table~\ref{tab:significance}.}
  \label{tab:pooled}
  \centering
  \input{tables/pooled}
\end{table}

\begin{figure}[t]
  \centering
  \includegraphics[width=\columnwidth]{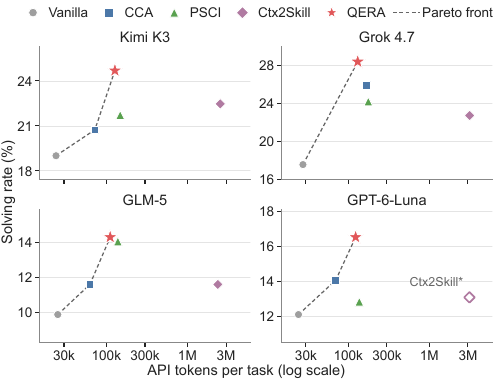}
  \caption{Solving rate against API tokens per task (log scale) on each LLM; the dashed line is the accuracy--cost Pareto front. Ctx2Skill$^{*}$ on GPT-6-Luna is drawn at its full cost: GPT-6-Luna inference plus the Grok 4.7 self-play whose skills it reuses.}
  \Description{Four scatter plots, one per LLM, of solving rate versus tokens per task. QERA, marked by a star, is the highest point in every panel and lies on the Pareto front; Ctx2Skill is far to the right at about three million tokens.}
  \label{fig:cost}
\end{figure}

\paragraph{\qera{} solves the most tasks on every LLM}
\qera{} attains the highest solving rate on all four LLMs and the highest average in every context category (Table~\ref{tab:main}). Averaged over the LLMs, it solves 21.0\% of the tasks, against 18.2\% for the strongest baseline (PSCI) and 14.6\% for Vanilla, a relative gain of 15\% and 43\%, and it passes 75.6\% of the rubrics, against 73.5\% (CCA) and 69.9\%. In the average block of Table~\ref{tab:main}, \qera{} is first in all eight columns, and in the per-LLM blocks it is first in 24 of the 32 columns. Pooled over the 1,620 task--LLM pairs, its lead is significant against every baseline in both metrics (Table~\ref{tab:pooled}): by 2.8--3.5 points of solving rate and 2.1--3.1 points of rubric pass rate over the three inference-time methods, and by 6.4 and 5.6 points over Vanilla.

\paragraph{No baseline is consistently second}
The strongest baseline changes with the LLM: by solving rate, it is Ctx2Skill on Kimi K3, CCA on Grok 4.7 and GPT-6-Luna, and PSCI on GLM-5, and each baseline's rank moves between second and fourth. PSCI even falls below Vanilla in rubric pass rate on GPT-6-Luna (66.8 vs.\ 67.1). \qera{} needs no such choice: it ranks first in solving rate on all four.

\paragraph{Per-LLM comparisons}
Per LLM, \qera{} is significantly ahead in 11 of the 16 rubric comparisons with a baseline and in 6 of the 16 solving comparisons (Table~\ref{tab:significance}), and behind in one: on Grok 4.7, CCA passes more rubrics ($-2.1$ [$-3.8$, $-0.4$]) but solves fewer tasks (25.9\% vs.\ 28.4\%). CCA is unusually strong on Grok 4.7: its gain over Vanilla is 6.8 rubric points there and 2--3 points on the other LLMs, and its Grok 4.7 replies are long (median 11.1k characters, 2.4 times \qera{}'s). Two thirds of the gap (1.4 of 2.1 points) comes from the 67 tasks on which CCA's reply is more than four times longer than \qera{}'s, so CCA's rubric lead on Grok 4.7 rests mainly on much longer replies. \qera{}'s own lead is not a matter of length: pooled over the LLMs, its median reply (4.7k characters) is shorter than CCA's (6.0k) and PSCI's (4.8k).

\paragraph{\qera{} is on the accuracy--cost front}
Figure~\ref{fig:cost} places each method by tokens and solving rate. \qera{} improves on PSCI in both metrics on every LLM while using 10--26\% fewer tokens (112k--131k vs.\ 137k--177k per task) and fewer calls (3.8--4.0 vs.\ 5). Ctx2Skill spends 2.4M--3.1M tokens and over 150 calls per task, nearly all in a self-play whose skill then serves a single task. \qera{} uses 4--5\% of these tokens and solves more tasks on every LLM. CCA is cheaper than \qera{} on three LLMs (63k--72k tokens) and solves fewer tasks on all four. Over Vanilla, \qera{}'s roughly fivefold token cost buys a 43\% relative gain in solving rate (Appendix~\ref{app:cost}).

\smallskip\noindent\textbf{Answer to RQ1.} Asking before answering improves real-life context learning on all four LLMs: \qera{} solves the most tasks on each of them, is significantly ahead of every baseline when pooled, and is cheaper than PSCI and Ctx2Skill.

\subsection{RQ2: Asking Before vs.\ Checking After}
\label{sec:rq2}

\begin{table}[t]
  \caption{Cumulative ablation on GPT-6-Luna (405 tasks). Impl.: pass rate on implicit rubrics. $\Delta$Rub.: paired difference of unrounded rates to the row above ($^{\dagger}$: to \qera{}), with 95\% bootstrap interval. Tok.: API tokens per task of the calls the configuration makes. The fourth row is \qera{}'s draft (details in Appendix~\ref{app:draft}).}
  \label{tab:ablation}
  \centering
  \input{tables/ablation}
\end{table}

\begin{figure}[t]
  \centering
  \includegraphics[width=\columnwidth]{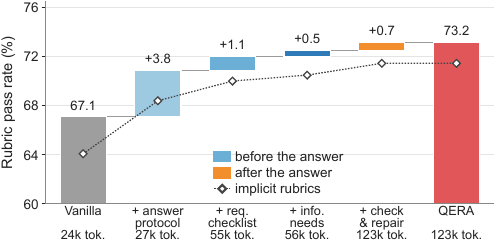}
  \caption{Rubric pass rate on GPT-6-Luna as \qera{}'s stages are added, with tokens per task. Blue steps act before the answer and the orange step after it; diamonds give the pass rate on implicit rubrics.}
  \Description{A waterfall chart from Vanilla at 67.1 to QERA at 73.2. The answer protocol adds 3.8 points, the checklist 1.1, the information needs 0.5 and check and repair 0.7; the token cost rises from 24k to 123k, mostly in the last step.}
  \label{fig:ablation}
\end{figure}

We build \qera{} up from Vanilla on GPT-6-Luna one stage at a time (Table~\ref{tab:ablation}, Figure~\ref{fig:ablation}): the answer protocol alone (one call), then the requirement checklist, then the information needs (the complete Ask stage, i.e., \qera{}'s draft), and finally check and repair (\qera{}). Each configuration re-runs only the calls its change affects; identical requests are served from the cache of the full run, so consecutive rows differ only in the added stage. For cost, this build-up runs on one LLM; the effect of checking is measured on all four. We also replace the check by a generic self-refinement that rereads the request and the materials, remove the measured facts and the audit, and remove the guard.

\paragraph{Asking before the answer produces most of the gain}
The answer protocol alone raises the solving rate from 12.1\% to 17.0\% (+4.9 [+1.7, +8.1]) and the rubric pass rate from 67.1\% to 70.9\% (+3.8 [+2.0, +5.7]) for 3k extra tokens per task. Over Vanilla with the same answer protocol, the request-specific questions add 1.6 rubric points [+0.1, +3.2] (checklist +1.1, information needs +0.5) and raise the solving rate from 17.0\% to 18.8\%, for one more call. Along this build-up order, the steps before the answer account for 5.4 of \qera{}'s 6.1-point rubric gain over Vanilla on GPT-6-Luna, and for 6.4 of its 7.4-point gain on implicit rubrics, whereas the pass rate on explicit rubrics stays between 80.2\% and 81.5\% across the five cumulative rows.

\paragraph{Checking after the answer refines numbers}
The check helps where an itemized second look with quotes can catch errors: on rubrics that mention numbers or counts (47\% of all rubrics), pooled over the LLMs, check and repair add 0.9 points [+0.1, +1.8], and they leave the other rubrics unchanged ($-0.3$ [$-0.9$, +0.3]). Overall, they move the rubric pass rate by +0.7 points on GPT-6-Luna and by +0.4, 0.0 and +0.1 on Kimi K3, Grok 4.7 and GLM-5, and the pooled solving rate by $-0.6$ [$-1.9$, +0.7], none of these changes significant, while they account for 51--56\% of \qera{}'s tokens (Appendix~\ref{app:draft}); a generic self-refinement~\cite{madaan2023selfrefine} reaches a similar rubric pass rate on GPT-6-Luna (73.2\%) with 82k instead of 123k tokens. The guard rejects 27 of the 1,385 repairs, mostly patches that would drop most of the draft's specifics or come back empty or cut off. The accuracy of \qera{} is thus largely in place before the answer: its two-call draft (54k--64k tokens) already solves more tasks than every baseline on every LLM with fewer tokens than CCA, PSCI or Ctx2Skill (Appendix~\ref{app:draft}), which makes the check an optional refinement where cost matters. This agrees with the observation that LLMs struggle to correct their own reasoning without external feedback~\cite{huang2024cannot}: asked before the answer, the questions change what the model reads for and writes.

\smallskip\noindent\textbf{Answer to RQ2.} The gain is settled before answering: the fixed answer protocol (+3.8 rubric points) and the request-specific questions of the Ask stage (+1.6) deliver 89\% of \qera{}'s rubric gain on GPT-6-Luna, and the draft alone solves more tasks than every baseline; checking after the answer refines numeric requirements.

\subsection{RQ3: Where Do the Gains Come From?}
\label{sec:rq3}

\begin{figure}[t]
  \centering
  \includegraphics[width=\columnwidth]{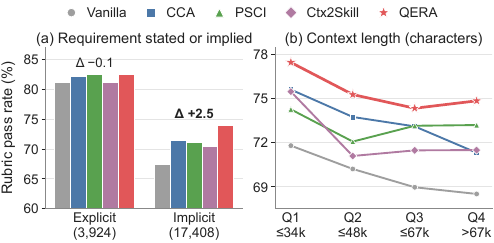}
  \caption{(a)~Rubric pass rate on explicit and implicit rubrics, pooled over the four LLMs (rubric--LLM pairs in parentheses), with \qera{} minus the best baseline. (b)~Rubric pass rate by context-length quartile (characters).}
  \Description{Left: grouped bars; on explicit rubrics all methods are near 82 percent, on implicit rubrics QERA reaches 74 percent, 2.5 points above the best baseline. Right: lines over four context-length quartiles; QERA is the top line in every quartile.}
  \label{fig:rq3}
\end{figure}

\paragraph{The gains are on implicit requirements}
On explicit rubrics, all five methods lie within 1.4 points of each other (81.1--82.5\%), and \qera{} ties the best baseline ($-0.1$ [$-1.9$, +1.6] against PSCI; Figure~\ref{fig:rq3}a). On implicit rubrics, which make up 82\% of the classified rubrics, \qera{} passes 74.0\%: 2.5 points [+1.4, +3.6] above the best baseline (CCA) and 6.6 points [+5.3, +7.9] above Vanilla. The baselines also gain mainly on implicit rubrics, but by less (3.1--4.1 points over Vanilla). The pattern holds at every level of overlap: on rubrics with under 20\%, 20--40\% and 40--60\% of their content words in the request, \qera{} leads the best baseline by 2.1, 2.7 and 2.3 points, each significant, and on rubrics with at least 60\% it ties. The questions help where the request's own words run out.

\paragraph{The gains hold across context lengths}
\qera{} is the best method in every context-length quartile, 1.2--1.8 points above the best baseline of that quartile (Figure~\ref{fig:rq3}b), which is CCA for the shorter half and PSCI for the longer half. All methods lose 1.1--4.3 points from the shortest to the longest quartile (\qera{} 2.6), in line with the benchmark's observation that difficulty is only weakly tied to length~\cite{dou2026clbenchlife}. Because \qera{} keeps the whole conversation in every call, its questions add focus without removing context.

\paragraph{The gains hold across context types}
\qera{} has the highest average in all three categories (Table~\ref{tab:main}). Its lead is largest in rubric pass rate in Communication \& Social Interactions (76.8\% vs.\ 74.7\% for CCA), where requests refer to people and turns of a conversation, and in solving rate in Fragmented Information \& Revisions (19.8\% vs.\ 16.5\%), where later versions supersede earlier ones; it is smallest in Behavioral Records \& Activity Trails, as discussed below. Across the 36 combinations of LLM and subcategory, \qera{} has the highest rubric pass rate in 20, against 8 for PSCI, 6 for CCA, 3 for Ctx2Skill and none for Vanilla (one tie; Table~\ref{tab:subcat}).

\paragraph{What the questions capture}
The Ask stage makes two kinds of implicit requirement explicit: rules from earlier turns, as in the chess task of \S\ref{sec:intro}, which the checklists of Kimi K3 and Grok 4.7 carry forward (both solve it, while Vanilla writes chess notation with all four LLMs), and complete enumerations: 47\% of the needs ask for every item that meets the user's criterion rather than the first ones found.

\paragraph{Where headroom remains}
Numbers and counts remain the hardest requirements: \qera{} passes 71.4\% of the rubrics that mention numbers or counts, against 79.2\% of the others (pooled over the LLMs). They make up 71\% of the rubrics in Behavioral Records \& Activity Trails (41\% and 32\% elsewhere), where \qera{}'s lead over PSCI is smallest (21.1\% vs.\ 20.7\% solved, 73.0\% vs.\ 72.4\% of rubrics).

\smallskip\noindent\textbf{Answer to RQ3.} \qera{}'s gains come from implicit rubrics, where it leads the best baseline by 2.5 points while tying on explicit ones, a lead that holds in every band of lexical overlap below 60\%; \qera{} ranks first in every context-length quartile and context category and in most LLM--subcategory combinations.

%% file: tables/main.tex
\small
\setlength{\tabcolsep}{4.6pt}
\begin{tabular}{@{}llrrrrrrrrr@{}}
\toprule
& & \multicolumn{2}{c}{Communication} & \multicolumn{2}{c}{Behavioral} & \multicolumn{2}{c}{Fragmented} & \multicolumn{2}{c}{Overall} & \\
\cmidrule(lr){3-4}\cmidrule(lr){5-6}\cmidrule(lr){7-8}\cmidrule(lr){9-10}
Model & Method & \makebox[2.9em][r]{Solv.} & \makebox[2.9em][r]{Rub.} & \makebox[2.9em][r]{Solv.} & \makebox[2.9em][r]{Rub.} & \makebox[2.9em][r]{Solv.} & \makebox[2.9em][r]{Rub.} & \makebox[2.9em][r]{Solv.} & \makebox[2.9em][r]{Rub.} & Tokens \\
\midrule
\multirow{5}{*}{Kimi K3} & Vanilla & 20.7 & 76.6 & 19.3 & 70.9 & 17.0 & 76.1 & 19.0 & 74.6 & 24k \\
 & CCA & 24.4 & 78.5 & 17.0 & 72.1 & \underline{20.7} & \underline{78.9} & 20.7 & 76.7 & 72k \\
 & PSCI & \textbf{28.1} & \underline{80.7} & \underline{23.7} & \underline{77.5} & 13.3 & 78.3 & 21.7 & \underline{78.9} & 148k \\
 & Ctx2Skill & \underline{26.7} & 79.4 & 20.7 & 74.4 & 20.0 & 76.8 & \underline{22.5} & 76.9 & 2.54M \\
 & \textbf{QERA} & \textbf{28.1} & \textbf{81.1} & \textbf{24.4} & \textbf{78.3} & \textbf{21.5} & \textbf{82.0} & \textbf{24.7} & \textbf{80.5} & 128k \\
\midrule
\multirow{5}{*}{Grok 4.7} & Vanilla & 18.5 & 74.8 & 17.8 & 72.4 & 16.3 & 74.7 & 17.5 & 74.0 & 28k \\
 & CCA & \underline{27.4} & \textbf{81.2} & 27.4 & \textbf{79.9} & \underline{23.0} & \textbf{81.1} & \underline{25.9} & \textbf{80.8} & 169k \\
 & PSCI & 20.7 & 76.4 & \textbf{31.1} & \underline{78.2} & 20.7 & 78.9 & 24.2 & 77.9 & 177k \\
 & Ctx2Skill & 23.7 & 79.0 & 23.7 & 78.0 & 20.7 & \underline{79.3} & 22.7 & \underline{78.8} & 3.14M \\
 & \textbf{QERA} & \textbf{28.1} & \underline{80.0} & \underline{30.4} & 77.5 & \textbf{26.7} & 78.6 & \textbf{28.4} & 78.7 & 131k \\
\midrule
\multirow{5}{*}{GLM-5} & Vanilla & 7.4 & 66.8 & 13.3 & 58.7 & 8.9 & 65.9 & 9.9 & 64.0 & 25k \\
 & CCA & 13.3 & 68.4 & 11.9 & 61.4 & 9.6 & \textbf{70.4} & 11.6 & 66.9 & 63k \\
 & PSCI & \textbf{16.3} & \underline{71.1} & \underline{14.1} & \textbf{67.0} & \underline{11.9} & 69.5 & \underline{14.1} & \underline{69.3} & 139k \\
 & Ctx2Skill & 14.1 & 68.4 & \underline{14.1} & 60.6 & 6.7 & 67.5 & 11.6 & 65.7 & 2.37M \\
 & \textbf{QERA} & \underline{14.8} & \textbf{72.4} & \textbf{14.8} & \underline{66.7} & \textbf{13.3} & \underline{70.1} & \textbf{14.3} & \textbf{69.8} & 112k \\
\midrule
\multirow{5}{*}{GPT-6-Luna} & Vanilla & 11.9 & 70.0 & \underline{14.1} & 63.2 & 10.4 & 67.8 & 12.1 & 67.1 & 24k \\
 & CCA & \underline{15.6} & \underline{70.5} & \underline{14.1} & 65.0 & \underline{12.6} & \underline{72.7} & \underline{14.1} & \underline{69.6} & 70k \\
 & PSCI & \underline{15.6} & 65.8 & \underline{14.1} & \underline{66.7} & 8.9 & 67.9 & 12.8 & 66.8 & 137k \\
 & Ctx2Skill$^{*}$ & 13.3 & 70.4 & \underline{14.1} & 64.7 & 11.9 & 70.1 & 13.1 & 68.5 & 3.13M \\
 & \textbf{QERA} & \textbf{17.0} & \textbf{74.0} & \textbf{14.8} & \textbf{69.7} & \textbf{17.8} & \textbf{75.5} & \textbf{16.5} & \textbf{73.2} & 123k \\
\midrule
\multirow{5}{*}{\begin{tabular}[c]{@{}l@{}}Average\\(4 models)\end{tabular}} & Vanilla & 14.6 & 72.0 & 16.1 & 66.3 & 13.1 & 71.1 & 14.6 & 69.9 & 25k \\
 & CCA & \underline{20.2} & \underline{74.7} & 17.6 & 69.6 & \underline{16.5} & \underline{75.8} & 18.1 & \underline{73.5} & 93k \\
 & PSCI & \underline{20.2} & 73.5 & \underline{20.7} & \underline{72.4} & 13.7 & 73.6 & \underline{18.2} & 73.2 & 150k \\
 & Ctx2Skill$^{*}$ & 19.4 & 74.3 & 18.1 & 69.4 & 14.8 & 73.4 & 17.5 & 72.5 & 2.80M \\
 & \textbf{QERA} & \textbf{22.0} & \textbf{76.8} & \textbf{21.1} & \textbf{73.0} & \textbf{19.8} & \textbf{76.5} & \textbf{21.0} & \textbf{75.6} & 123k \\
\bottomrule
\end{tabular}

%% file: tables/pooled.tex
\small
\begin{tabular*}{\columnwidth}{@{\extracolsep{\fill}}lll@{}}
\toprule
QERA $-$ & $\Delta$ Solving rate & $\Delta$ Rubric pass rate \\
\midrule
Vanilla & +6.4 {\scriptsize[+4.3, +8.5]} & +5.6 {\scriptsize[+4.5, +6.8]} \\
CCA & +2.9 {\scriptsize[+0.9, +4.9]} & +2.1 {\scriptsize[+1.1, +3.1]} \\
PSCI & +2.8 {\scriptsize[+0.7, +4.9]} & +2.4 {\scriptsize[+1.4, +3.3]} \\
Ctx2Skill & +3.5 {\scriptsize[+1.5, +5.6]} & +3.1 {\scriptsize[+2.2, +4.0]} \\
\bottomrule
\end{tabular*}

%% file: tables/ablation.tex
\footnotesize
\setlength{\tabcolsep}{3.2pt}
\begin{tabular}{@{}lrrrlr@{}}
\toprule
Configuration & Solv. & Rub. & Impl. & $\Delta$Rub. vs.\ row above & Tok. \\
\midrule
Vanilla & 12.1 & 67.1 & 64.1 & -- & 24k \\
\multicolumn{6}{@{}l}{\emph{Before the answer}} \\
+ answer protocol & 17.0 & 70.9 & 68.4 & +3.8 {\scriptsize[+2.0, +5.7]} & 27k \\
+ requirement checklist & 16.8 & 72.0 & 70.0 & +1.1 {\scriptsize[$-$0.3, +2.6]} & 55k \\
+ information needs (draft) & 18.8 & 72.5 & 70.5 & +0.5 {\scriptsize[$-$1.0, +2.0]} & 56k \\
\midrule
\multicolumn{6}{@{}l}{\emph{After the answer}} \\
+ check \& repair (\textbf{QERA}) & 16.5 & 73.2 & 71.5 & +0.7 {\scriptsize[$-$0.6, +2.0]} & 123k \\
\quad self-refine, no check & 18.5 & 73.2 & 71.0 & +0.1 {\scriptsize[$-$1.3, +1.4]}$^\dagger$ & 82k \\
\quad check w/o code facts & 17.8 & 73.1 & 71.2 & 0.0 {\scriptsize[$-$1.3, +1.3]}$^\dagger$ & 122k \\
\quad no guard & 16.5 & 73.2 & 71.5 & 0.0 {\scriptsize[0.0, +0.1]}$^\dagger$ & 123k \\
\bottomrule
\end{tabular}

%% file: sections/conclusion.tex
\section{Conclusion and Limitations}
\label{sec:conclusion}

On real-life context, the request is not the specification: most of what a correct reply must contain is implicit in the conversation. \qera{} recovers it by asking before answering, with a requirement checklist and typed information needs kept faithful by code, and asks the same questions again with verified evidence. On all of CL-bench Life and four frontier LLMs, it solves the most tasks of the inference-time methods we compared, at a moderate cost (RQ1); its gain is settled before the answer, by a fixed answer protocol and the request-specific questions, rather than by the check after it (RQ2); and the gain lies on implicit requirements, across context lengths and context types (RQ3). For assistants that work over users' web traces, inference is best spent on recovering what a request leaves unsaid before generating.

\paragraph{Limitations}
While \qera{} achieves consistent gains across LLMs and context types, we see several directions for future exploration. First, \qera{} is built on LLMs throughout: the same LLM writes the requirement checklist and the information needs, answers them and checks its own answer, and an LLM judge grades the result. Code keeps the questions faithful to the user's words and verifies every quote, yet the underlying LLM can still hallucinate or misread messy context, for example by missing a rule stated turns earlier or by misjudging which records meet a criterion, and such errors carry into the answer. The quality of the questions may therefore depend on the capabilities of the underlying LLM, and different LLMs may ask somewhat different questions; stronger LLMs, or questions verified against the context, are likely to widen the gains. The judge, too, is an LLM: the benchmark validated it against human checks and other judges (\S\ref{sec:setup}), but individual verdicts can still be wrong. Second, asking and checking cost tokens: \qera{} uses about five times the tokens of direct answering, although fewer than PSCI and only 4--5\% of Ctx2Skill's. Because the gain is settled before the answer, where \qera{}'s draft alone already solves more tasks than every baseline, a check that runs only when the questions flag a risk could cut this cost further. Finally, our study covers text: chats, posts, logs and documents. Real-life traces also include photos, screenshots, voice messages and videos; extending \qera{}'s questions to such multimodal context is a natural next step.

\paragraph{Ethics}
We use CL-bench Life under its license, for evaluation only, and collected no new data; \qera{} processes a conversation only at inference time and stores nothing. Each stage sends the conversation to the LLM provider, so a deployment inherits the provider's data-retention terms, which locally hosted LLMs avoid. Assistants that reason over personal traces raise questions of consent and data governance, which remain with whoever deploys them.

%% file: sections/appendix.tex
\section{Prompts}
\label{app:prompts}
\begin{table*}[t]
  \caption{Results by subcategory (45 tasks each), averaged over the four LLMs. Bold: best; underlined: second best.}
  \label{tab:subcat}
  \centering
  \input{tables/subcat}
\end{table*}
\begin{table*}[t]
  \caption{\qera{} minus each baseline per LLM and pooled, in percentage points of unrounded rates, with paired 95\% bootstrap intervals over tasks; grey: the interval contains zero. Ctx2Skill on GPT-6-Luna is Ctx2Skill$^{*}$.}
  \label{tab:significance}
  \centering
  \input{tables/significance}
\end{table*}
To support reproduction, Appendices~\ref{app:prompts}--\ref{app:baselines} and \S\ref{sec:setup} give the prompts, every deterministic rule and threshold, and all model, baseline and judge settings.
\input{sections/appendix_prompts}

\section{Filters, Audit and Guard}
\label{app:filters}

All rules are code; ``user's words'' means all user turns, matched after normalization. A user message over 4,000 characters counts as pasted materials, and only its last 2,000 characters are read as the request (1 of the 405 tasks); length cues are read in the request and in earlier user turns under 2,000 characters.
\emph{Filters $\Phi$ on the checklist.}
(1)~Exclusions are dropped when their grounding quote is not in the user's words, when a purpose-implied exclusion states no purpose, when they are caution instructions, or when they exclude missing values although the user asked to mark unknowns.
(2)~Caution clauses (``do not infer/assume/overstate'', ``as an established fact'', ``state the limitation'', ``cautiously'', denials of proof) are cut from requirements, scopes and readings unless the user's words contain them.
(3)~Brevity words (concise, brief, succinct, ``short'' before a reply noun, ``keep it brief'') are cut when no user turn contains a length cue.
(4)~If the request names no time, a date window in a scope clause about all of the data is cut when it is narrower than the materials' span of dates.
(5)~An earlier-reply template is dropped when that reply is missing or contains none of its headings or entities, or when the last message limits the form (``just list'', ``nothing else'').
(6)~Required strings absent from the user's words are dropped; the rest take the user's spelling.
(7)~An unrequested JSON, YAML or XML format is reset to ``not specified''.
\emph{Filters on the needs.}
(8)~Rule (2) applies to needs, and narrowing words the user never used (sustained, explicitly stated, confirmed, significant) are cut. (9)~Date windows whose months never occur in the conversation are cut from a need and its ``where''; rule (4) applies too. (10)~\emph{Coverage}: a list, count, timeline, summary or classification requirement that no enumerate, count or compute need serves gets a first need for ``every item, unit or event the user's words cover'' for it, each with its identifier as the materials write it.
\emph{Actionable check and audit.}
(11)~Only items with something to act on reach the repair; a verdict alone never triggers one. (12)~Findings and unsupported-claim items whose quote (12 or more characters) is not found in the conversation are marked ``[quote not found]''; items relying on a year outside the conversation's years are dropped. (13)~Using only findings whose quote is found (or that have none), the audit reports the code's sum when at least three listed inputs of a compute need miss its stated result (as a sum and as a mean, 0.5\% tolerance) and the draft lacks that sum; a count that differs from the number of distinct listed items; for an enumerate or count need marked correct that serves a list requirement, the found items that the draft never names, when they are at most 15 and at most half of the items; and up to ten numbers or dates found for an item the draft names but whose value it lacks. A repair is called when any such line or actionable item remains, or when a required string is missing, an excluded item is mentioned or a label leaks.
\emph{Guard $\Gamma$.} The repair is rejected, and the draft kept, when it is (14)~empty, or cut off mid-sentence while the draft is not; (15)~under half of the draft unless a length fix was asked or the draft broke the user's limits; (16)~adds a year outside the conversation's range; (17)~keeps under 30\% of the draft's numbers and times when the draft has ten or more (with the same exceptions as (15)); (18)~exceeds a whole-reply word limit (of at least 50 words; 10\% tolerance unless stated as a maximum), or breaks a sentence, bullet or paragraph count, that the draft met; (19)~is over 1.25 times a draft the user asked to keep brief while nothing failed; or (20)~drops a required string that the draft contained, or adds a leaked label or an excluded item. A final polish restores quotations to their source text, sets required strings to the user's spelling, removes HTML tags from tables, and removes bold and italics when a user turn bans them.

\section{Baseline Implementation}
\label{app:baselines}

\emph{Shared protocol.} Every method uses one frozen LLM for all of its calls (except Ctx2Skill$^{*}$), high reasoning effort and the model's full output limit (128K tokens for Kimi K3 and GPT-6-Luna; the rest of the 524,288- and 202,752-token windows for Grok 4.7 and GLM-5); no temperature or seed is set, as the reasoning models fix their sampling. All methods receive the original conversation, and none observes rubrics, reference answers or other CL-bench Life tasks. The judge is GPT-5.1 (snapshot 2025-11-13, high reasoning, one user message, no temperature or output cap) with the benchmark's grading template, whose first sentence ``Starting now, you are a rigorous instruction-following grading teacher.'' is replaced by ``Assess the response according to the evaluation criteria.''; the rest of the template, the rubric list, the separators and the output format are unchanged. With the original sentence, requests returned HTTP 400 (\texttt{invalid\_prompt}, flagged as a possible usage-policy violation) in our environment, also for benign synthetic inputs, and replacing that sentence alone restored them. Before grading real answers, we checked that the adapted template returns the expected per-rubric verdicts on four synthetic fixtures. Every method is graded with the same request builder and parser. Infrastructure failures, including GPT-6-Luna's occasional refusal of a harmless prompt as a policy violation, are retried for every method, and no output is filtered or selected; requests to GPT-6-Luna that Azure's content filter refused were sent to OpenAI's own API. Four of the 8,100 task runs still ended without an answer and count as failed: one Vanilla and two PSCI tasks on GPT-6-Luna, refused on every retry, and one Ctx2Skill task on GLM-5, which returned reasoning but no text. Model identifiers: \texttt{global.moonshotai.kimi-k3}, \texttt{global.xai.grok-4.7} and \texttt{zai.glm-5} on Amazon Bedrock (for Grok 4.7, \qera{}'s runs fall back to the profile \texttt{us.xai.grok-4.7} of the same model when the first is throttled), \texttt{gpt-6-luna} version 2026-09-22 on Azure OpenAI, and the judge \texttt{gpt-5.1-2025-11-13}.
\emph{CCA~\cite{qi2026cca}.} The released pipeline (Compiler, CodeGen, Reasoner, verifier execution, and a second Reasoner pass when the verifiers flag two or more violations) runs with unchanged prompts, schema and correction logic; only the model-call layer is replaced. CL-bench Life tasks have no system prompt, so an empty one is prepended (CCA reads the second message as the context), and the context caps are raised to 600K characters so that nothing is truncated. The Compiler never failed, so the fallback to direct prompting was never used.
\emph{PSCI~\cite{zhong2026psci}.} We use the paper's five prompt templates verbatim (source-only and rotated-demonstration contract induction, contract-guided generation, adversarial checking, one repair) and its pipeline. Unspecified details: ten rotated demonstrations per task, each the rubric list of one CL-bench task (never CL-bench Life), seeded by the task id; contracts merged with source-only items first and duplicates removed; at most eight repairs; a stage whose JSON does not parse is asked once more, after which a failed induction contributes nothing and a failed check keeps the draft. The templates take the source in one slot, so the conversation is passed as one message with its turns in order and their roles marked.
\emph{Ctx2Skill~\cite{si2026ctx2skill}.} The released code runs unchanged (five self-play iterations with five probing tasks per context, Cross-Time Replay, the official inference script), once for each of the 404 contexts. The paper fixes the self-play judge to GPT-5.1; we let the evaluated LLM play all five roles, so that no method receives supervision from a stronger model, and use high effort for every role (the code sends no reasoning setting). It makes 153--163 calls and uses 2.4M--3.1M tokens per task, 87\% of the tokens of all five methods on these three LLMs. On GPT-6-Luna we therefore reuse, per context, the Reasoner skill that Cross-Time Replay selected in the Grok 4.7 self-play, Ctx2Skill's strongest run (22.7\% solved, 78.8\% rubric pass rate), and run the official inference on GPT-6-Luna; a usable skill is found for 393 of the 404 contexts, as in the Grok 4.7 run. This favours Ctx2Skill$^{*}$, whose skills come from a stronger self-play backbone. We charge it the full cost, the reused self-play (3.10M Grok 4.7 tokens and 152 calls per task) plus GPT-6-Luna's inference (28k tokens, one call); Tables~\ref{tab:main} and~\ref{tab:cost} and Figure~\ref{fig:cost} all use this total.

\section{Cost Breakdown}
\label{app:cost}

\begin{table}[tb]
  \caption{API tokens (input + output, reasoning included) and LLM calls per task. $^{*}$GPT-6-Luna inference plus the reused Grok 4.7 skill discovery.}
  \label{tab:cost}
  \centering
  \input{tables/cost}
\end{table}

Table~\ref{tab:cost} lists tokens and calls per task. \qera{}'s tokens split into Ask 21--23\%, Answer 22--27\%, Check 29--35\% and Repair 17--23\%; every call re-reads the conversation (22k--25k input tokens), so input dominates (65--79\%). The repair runs on 76--95\% of tasks; JSON repairs occur only on GLM-5 (0.08 calls per task). Tokenizers differ, so compare tokens within an LLM. \qera{}'s calls run one after another, so its latency is that of three or four calls of the same LLM; its code steps take milliseconds. We do not report wall-clock times, which in our runs were dominated by provider rate limits under high concurrency.

\section{Draft vs.\ Final}
\label{app:draft}

\begin{table}[tb]
  \caption{\qera{}'s draft (Ask + Answer) vs.\ final reply: differences of unrounded rates with paired 95\% intervals; tokens per task (draft/all); $^{\ddagger}$repairs kept/guard-rejected/invalid/nothing to fix.}
  \label{tab:draft}
  \centering
  \scriptsize
  \setlength{\tabcolsep}{1.6pt}
  \begin{tabular}{@{}lccllrr@{}}
    \toprule
    LLM & Draft S/R & Final S/R & $\Delta$Solv. & $\Delta$Rub. & Tokens & Repairs$^{\ddagger}$ \\
    \midrule
    Kimi K3    & 25.2/80.2 & 24.7/80.5 & $-$0.5 [$-$3.0, +2.0] & +0.4 [$-$0.9, +1.4] & 56k/128k & 334/15/8/48 \\
    Grok 4.7   & 27.7/78.8 & 28.4/78.7 & +0.7 [$-$1.7, +3.2] & 0.0 [$-$0.9, +0.8] & 64k/131k & 307/0/0/98 \\
    GLM-5      & 14.6/69.7 & 14.3/69.8 & $-$0.2 [$-$2.7, +2.2] & +0.1 [$-$1.0, +1.2] & 54k/112k & 336/2/0/67 \\
    GPT-6-Luna & 18.8/72.5 & 16.5/73.2 & $-$2.2 [$-$5.4, +0.7] & +0.7 [$-$0.6, +2.0] & 56k/123k & 381/2/0/19 \\
    \midrule
    Pooled     & 21.5/75.3 & 21.0/75.6 & $-$0.6 [$-$1.9, +0.7] & +0.3 [$-$0.3, +0.8] & 57k/123k & 1358/19/8/232 \\
    \bottomrule
  \end{tabular}
\end{table}

In Table~\ref{tab:ablation}, one task in each of the first three rows has no answer (refused by GPT-6-Luna on every retry), and row 2's answers still pass through the code polish, which restores the spelling of required strings from the cached checklist (18 of 404 answers). Per Table~\ref{tab:draft}, the draft alone solves more tasks than every baseline on every LLM (best baselines: 22.5, 25.9, 14.1 and 14.1\%) with fewer tokens than CCA, PSCI, and Ctx2Skill counting its self-play; its rubric pass rate trails CCA's on Grok 4.7 only. Check and repair raise the pass rate on rubrics about numbers (2,513 rubrics; pooled +0.9 [+0.1, +1.8]) and leave the other rubrics unchanged ($-0.3$ [$-0.9$, +0.3]). Typical errors of the check and repair: the check can call knowledge the user asked for unsupported (on one Grok 4.7 task, correct titles of the next books in a series were replaced by placeholders: 15 to 5 of 32 rubrics); the guard does not detect garbled text (two Kimi K3 repairs: 22 and 9 rubrics to none); and a per-item length limit can be read as a limit on the whole reply (``a high level categorization and brief 1-2 sentence summary of the type of topics'', Grok 4.7: 759 characters, 1 of 13 rubrics; CCA: 12).

\section{Full Results}
\label{app:full}

\paragraph{Explicit and implicit rubrics}
A rubric's content words are its quoted strings, capitalized names, numbers and lowercase words of five or more letters (at most twelve), after removing a trailing ``For example \ldots'' clause and boilerplate (``the response should'', include, mention, \ldots). A rubric is explicit when at least 60\% of them occur in the request (the last user message, or its part after the pasted materials): 981 rubrics are explicit, 4,352 implicit, and 15 with fewer than two content words are left out.

By subcategory (Table~\ref{tab:subcat}), \qera{} has the highest solving rate in seven of nine (two tied) and the highest rubric pass rate in seven; Table~\ref{tab:significance} gives the per-LLM tests of \S\ref{sec:rq1}.

%% file: tables/subcat.tex
\footnotesize
\setlength{\tabcolsep}{3.2pt}
\begin{tabular}{@{}lrrrrrrrrrrc@{}}
\toprule
& \multicolumn{5}{c}{Solving rate (\%)} & \multicolumn{5}{c}{Rubric pass rate (\%)} & Best Rub. \\
\cmidrule(lr){2-6}\cmidrule(lr){7-11}\cmidrule(l){12-12}
Subcategory & Vanilla & CCA & PSCI & Ctx2Skill & QERA & Vanilla & CCA & PSCI & Ctx2Skill & QERA & K\,/\,G\,/\,Z\,/\,L \\
\midrule
\multicolumn{12}{@{}l}{\emph{Communication \& Social Interactions}} \\
\quad Community Interactions & 12.2 & 15.6 & \textbf{18.9} & \underline{17.8} & 17.2 & 71.5 & 74.3 & \textbf{75.6} & 72.9 & \underline{74.6} & P\,/\,C\,/\,\textbf{Q}\,/\,S \\
\quad Group Conversations \& Meeting Transcripts & 19.4 & 23.3 & 21.7 & \underline{23.9} & \textbf{27.2} & 76.5 & 78.7 & 75.4 & \underline{78.9} & \textbf{81.2} & \textbf{Q}\,/\,S\,/\,\textbf{Q}\,/\,\textbf{Q} \\
\quad Private Conversations & 12.2 & \textbf{21.7} & \underline{20.0} & 16.7 & \textbf{21.7} & 67.7 & 70.7 & 69.0 & \underline{71.0} & \textbf{74.8} & \textbf{Q}\,/\,C\,/\,P\,/\,\textbf{Q} \\
\midrule
\multicolumn{12}{@{}l}{\emph{Behavioral Records \& Activity Trails}} \\
\quad Digital Footprints \& Daily-Life Records & 11.7 & 9.4 & \underline{17.2} & \textbf{17.8} & \textbf{17.8} & 63.9 & 68.0 & \underline{71.2} & 68.6 & \textbf{71.5} & P\,/\,P\,/\,P\,/\,\textbf{Q} \\
\quad Game Logs & 24.4 & 29.4 & \textbf{30.6} & 26.1 & \underline{30.0} & 74.3 & 76.3 & \underline{79.8} & 76.6 & \textbf{80.0} & \textbf{Q}\,/\,C\,/\,P\,/\,P=Q \\
\quad Self-Tracking Trajectories & 12.2 & 13.9 & \underline{14.4} & 10.6 & \textbf{15.6} & 58.4 & 62.7 & \underline{64.1} & 61.2 & \textbf{65.7} & \textbf{Q}\,/\,C\,/\,\textbf{Q}\,/\,\textbf{Q} \\
\midrule
\multicolumn{12}{@{}l}{\emph{Fragmented Information \& Revisions}} \\
\quad Creation \& Revision Histories & 14.4 & \underline{19.4} & 17.8 & 17.2 & \textbf{22.2} & 69.9 & \underline{73.7} & 73.1 & 73.2 & \textbf{76.4} & \textbf{Q}\,/\,\textbf{Q}\,/\,\textbf{Q}\,/\,\textbf{Q} \\
\quad Personal Information Fragments & 14.4 & \underline{18.9} & 14.4 & 15.6 & \textbf{23.3} & 70.8 & \underline{75.4} & 72.6 & 72.8 & \textbf{76.6} & \textbf{Q}\,/\,S\,/\,P\,/\,\textbf{Q} \\
\quad Public Information Fragments & 10.6 & 11.1 & 8.9 & \underline{11.7} & \textbf{13.9} & 72.2 & \textbf{77.4} & 74.9 & 74.1 & \underline{76.6} & \textbf{Q}\,/\,C\,/\,C\,/\,\textbf{Q} \\
\bottomrule
\end{tabular}

%% file: tables/significance.tex
\footnotesize
\setlength{\tabcolsep}{3pt}
\begin{tabular}{@{}lllllll@{}}
\toprule
QERA $-$ & & Kimi K3 & Grok 4.7 & GLM-5 & GPT-6-Luna & Pooled \\
\midrule
Vanilla & Solv. & +5.7 {\scriptsize[+1.5, +9.9]} & +10.9 {\scriptsize[+6.7, +15.3]} & +4.4 {\scriptsize[+1.2, +7.9]} & +4.4 {\scriptsize[+1.0, +7.9]} & +6.4 {\scriptsize[+4.3, +8.5]} \\
 & Rub. & +5.9 {\scriptsize[+4.0, +7.7]} & +4.7 {\scriptsize[+3.0, +6.5]} & +5.8 {\scriptsize[+3.8, +7.9]} & +6.1 {\scriptsize[+3.9, +8.3]} & +5.6 {\scriptsize[+4.5, +6.8]} \\
\addlinespace[2pt]
CCA & Solv. & \textcolor{gray}{+4.0 {\scriptsize[0.0, +7.9]}} & \textcolor{gray}{+2.5 {\scriptsize[$-$1.7, +6.7]}} & \textcolor{gray}{+2.7 {\scriptsize[$-$0.7, +6.2]}} & \textcolor{gray}{+2.5 {\scriptsize[$-$1.5, +6.4]}} & +2.9 {\scriptsize[+0.9, +4.9]} \\
 & Rub. & +3.9 {\scriptsize[+2.0, +5.7]} & $-$2.1 {\scriptsize[$-$3.8, $-$0.4]} & +2.9 {\scriptsize[+0.8, +4.9]} & +3.6 {\scriptsize[+1.7, +5.6]} & +2.1 {\scriptsize[+1.1, +3.1]} \\
\addlinespace[2pt]
PSCI & Solv. & \textcolor{gray}{+3.0 {\scriptsize[$-$1.2, +7.2]}} & +4.2 {\scriptsize[+0.2, +8.1]} & \textcolor{gray}{+0.2 {\scriptsize[$-$3.2, +3.7]}} & \textcolor{gray}{+3.7 {\scriptsize[0.0, +7.4]}} & +2.8 {\scriptsize[+0.7, +4.9]} \\
 & Rub. & \textcolor{gray}{+1.7 {\scriptsize[$-$0.2, +3.4]}} & \textcolor{gray}{+0.9 {\scriptsize[$-$0.9, +2.6]}} & \textcolor{gray}{+0.5 {\scriptsize[$-$1.5, +2.5]}} & +6.4 {\scriptsize[+4.3, +8.7]} & +2.4 {\scriptsize[+1.4, +3.3]} \\
\addlinespace[2pt]
Ctx2Skill & Solv. & \textcolor{gray}{+2.2 {\scriptsize[$-$2.2, +6.7]}} & +5.7 {\scriptsize[+1.5, +9.9]} & \textcolor{gray}{+2.7 {\scriptsize[$-$0.2, +5.9]}} & \textcolor{gray}{+3.5 {\scriptsize[$-$0.2, +7.4]}} & +3.5 {\scriptsize[+1.5, +5.6]} \\
 & Rub. & +3.6 {\scriptsize[+1.8, +5.4]} & \textcolor{gray}{$-$0.1 {\scriptsize[$-$1.8, +1.5]}} & +4.1 {\scriptsize[+2.2, +6.0]} & +4.7 {\scriptsize[+2.8, +6.6]} & +3.1 {\scriptsize[+2.2, +4.0]} \\
\bottomrule
\end{tabular}

%% file: sections/appendix_prompts.tex
\scriptsize
Every call starts with the whole conversation. We give the answer protocol and the repair rules verbatim and summarize the other prompts closely.
\paragraph{Ask (one JSON call)}
An analyst instruction (``You are a meticulous research analyst working behind the scenes for the assistant in the conversation above \ldots'') and a closed-world rule for facts precede two steps. \emph{Step 1, requirement checklist}: ``Identify everything a complete, excellent reply to the user's LAST message must deliver.''
\begin{itemize}[leftmargin=1.2em,itemsep=0pt,topsep=1pt,parsep=0pt]
\item Split the request into separate requirements, one per distinct thing asked; for each give the user's exact words that create it, a precise restatement, its scope (people, items, records, period, units, and any criterion or definition the user gave) and its kind.
\item User-defined rules (a term, filter, threshold, formula, label, fallback or condition) are copied word for word into scope and never widened, narrowed or re-read; undefined ordinary words keep their ordinary reach, and no unstated condition (only from a source, explicitly stated, at least N) may remove items. Readings only for genuinely ambiguous wording, the everyday reading first.
\item Requirements from earlier user turns only if the last message relies on them; format, style and language rules, definitions and splits set earlier still apply unless replaced. When the last message edits or refers to an earlier reply, a \emph{preserve} requirement copies that reply's headings, item skeleton and entities.
\item Small list, count or compute requests (up to about 25 items) also require each item with its identifier and the attribute asked about; item\_fields list every attribute asked for per item. Each yes/no question gets a requirement to open with a direct verdict; inferences, judgments and classifications require a committed answer; an artifact request requires the artifact; the user's premise about their own situation is accepted unless the materials contradict it.
\item No uncertainty, caveat, verification or confidence instructions unless the user asked for them; a filter excludes an item only when the materials state a value that fails it. Exclusions need the user's exact words or a stated purpose with evidence in the materials; a criterion for which items qualify is scope, not an exclusion.
\item Required literals are only strings the user requires verbatim; the answer format and its source (user-stated, earlier reply, implied, not specified) are recorded, length or tone only if the user stated them. Do not answer yet; never drop a requirement because the materials might not support it.
\end{itemize}
\emph{Step 2, information needs}: ``List the facts a correct, complete reply must find or derive in the materials, so that a checker can later search the whole conversation for each one and compare it with a draft reply.''
\begin{itemize}[leftmargin=1.2em,itemsep=0pt,topsep=1pt,parsep=0pt]
\item One need per distinct thing to locate or derive, in the user's terms and criterion words; requirement\_ids name the requirements it serves; where = the document, thread, message, section or speaker to start from (dates only as the materials print them). Kinds: lookup, enumerate (every item meeting the criterion; the reply must name each), count, compute (including the user's own figures), compare, judge.
\item A list, count, summarize or classify requirement first gets one enumerate or count need with the user's criterion verbatim over all the materials (``every <unit>, with its identifier as the materials write it''); one enumerate need per category the user names; rates use the user's own unit and baseline; when the materials may not state what is asked, the need asks whether they do, never for a proxy.
\item Also needs for later corrections, updates or cancellations of a fact the reply uses, and for each figure the user's own numbers imply. About 3 to 25 needs; do not answer them.
\end{itemize}
\paragraph{Answer (one call): the answer protocol}
``Write your reply to the user's last message now.'', the filtered checklist and needs as advisory input (\S\ref{sec:answer}), then (verbatim): Before writing, read every part of the conversation that bears on each part of the request: every document, log, record and earlier message, not only the first match, and every later message, entry or version that corrects, updates, cancels or reverses an earlier one (state the current value; give the earlier one too when the user asks about change). For list, count or compute requests, go through the materials item by item and do the arithmetic step by step; state the results exactly.
\begin{itemize}[leftmargin=1.2em,itemsep=0pt,topsep=1pt,parsep=0pt]
\item Deliver every requirement with the specifics the materials give: names, titles and IDs as the materials write them, numbers with units, full dates and times, each party's position in their own key words, and the condition, reason or qualifier clauses; never replace a specific with a category word. A thorough, specific reply beats a vague one.
\item For itemized, list or per-item requests, include every qualifying item with its requested fields; brevity never drops them unless the user set a length or count limit. Keep record-level items at item granularity, in source order, unless the user asked for aggregation. Name records by the identifier the materials use (ID, title, full date) as well as any ordinal, and give the source of a claim when the user asks for evidence.
\item Yes/no, count, ranking and classification questions open with one sentence in the user's key words giving the verdict (Yes or No; ``Partly'' only naming which part is which), count, winner or class, except inside text the user asked you to write. Show each requested per-item value next to its item, also when a total is given. Use the format the user asked for (table, prose, article, letter, JSON, exact template), with the requested fields in its columns or sentences, and head the sections the user names with those names (spelling corrected).
\item Commit to definite answers: state counts, rankings and totals exactly, never ``about'', ``approximately'' or ``provisional''. For a top-N request always name N items (if there is a tie at the cutoff, name the tied items briefly). When asked to infer, judge, classify, estimate, choose or recommend, give the most likely answer plainly with its main evidence; do not add verdicts the user did not ask for.
\item Take the figures the user gives about their own situation (amounts, percentages, rates, dates) as stated unless the materials contradict them, and work out the concrete values they imply for each part of the answer (13\% of \$500 a month = \$65; the rest = \$435). For each total, difference, share, average or rate you report, give its inputs and operation in a short clause next to the result (8 / 145 = 5.5\%) unless a length limit leaves no room; count each event once even when several places mention it, on the basis the user's words name.
\item Include an item when it meets the user's criteria in their ordinary meaning; add no threshold, time window or evidence standard of your own, and never drop a qualifying item because a requested field is not stated (write ``\textless{}field\textgreater{}: not stated'').
\item When the checklist gives readings for a phrase, answer the first reading; give the result under another reading in one sentence only when it changes the answer (a different count, item set or verdict).
\item Say plainly, in one short clause where it first matters, the definition, criterion, time window or formula you applied when the question uses a term that needs one, and any value you inferred rather than read (a missing year) with the clue it rests on. Otherwise no disclaimers or process commentary: do not describe your research or how you resolved inconsistencies, and do not discuss data quality or what you can or cannot prove. Mention a limitation only when a requested item is entirely absent from the materials, in one short clause for that item.
\item Follow the requested format, structure, length and language; length and count limits the user set take precedence over completeness. When the user refers to an earlier reply's structure, reuse its headings and item layout, apply the user's changes, and keep its items unless the user removed them. Respect every exclusion: do not mention excluded items at all, not even to say they were left out, and never add a section, note or list of what you left out or why, unless a requirement asks about the exclusions.
\item Keep every attribution: each quote, claim and number names its speaker or source as the materials display it; never merge different people's claims. Quote whole sentences unless the user asks for short quotes or a length limit applies; never change words inside quotation marks; cut only whole unrelated sentences, marked with an ellipsis.
\item Outside any text the user asked you to write (story, letter, post, statement), speak as the assistant: the user is ``you'', and the participants in the materials are named in the third person. Inside such text, make the improvements directly: no drafting notes, editor's notes or remarks on how it differs from the original. No HTML tags in markdown tables.
\item When the materials show a timestamp as concatenated display variants (e.g. ``Today at 9:05 AMMonday, June 1, 2026 at 9:05 AM''), write the single absolute form; otherwise keep each timestamp in the source's format, and never turn a relative or weekday label into a calendar date unless the materials give that date. Decode HTML entities.
\item Facts about the specific people, events, records and works come only from the conversation; never add outside facts, mechanics or causes the materials do not support. General knowledge may fill only what the conversation does not define: calendar arithmetic, the standard meaning of a term or convention the user uses, and (only when the user asks to estimate, project or price something the materials do not give) a typical value labeled as an estimate; none when the user restricts the answer to the provided text. ``Now'' is the date the conversation states, else the date of the latest message or record entry that has already happened (never a scheduled or planned future date), never your own clock.
\item Never mention a checklist or this process. Output only the reply to the user.
\end{itemize}

\paragraph{Check (one JSON call)}
Given the checklist, the needs, the code-measured facts and the draft, the model checks the draft ``strictly, as a careful grader who knows only this conversation would'' and returns:
\begin{itemize}[leftmargin=1.2em,itemsep=0pt,topsep=1pt,parsep=0pt]
\item \emph{requirement\_checks}: per requirement, a concrete visible test in the user's words; \emph{pass} only with a short quote from the draft that satisfies it, otherwise \emph{partial} or \emph{fail} with the problem and the exact correction.
\item \emph{need\_checks}: per need, findings searched in the entire conversation (later corrections included), each with an exact quote under 25 words and its location. Enumerate and count needs go through the materials in order with one finding per qualifying item (identifier and value), never only the items the draft already has; count needs state how many items were listed, compute needs list each input and the result. Status \emph{correct} only when every finding is visible in the draft. A count or total is \emph{wrong} only when the problem names the item missed, double-counted or wrongly included, or the wrong input or step; a judgment only when a stated fact contradicts it.
\item \emph{latent\_gaps}: what the user's words in any turn, or a constraint stated earlier (a budget, diet, date, person, something already done), require of this answer that neither the checklist nor the draft covers; never a preference of the checker.
\item \emph{unsupported}: stated facts about the people, events or records that the materials contradict or never give, with the conflicting quote and the fix; a requested inference resting on stated evidence is not unsupported.
\item \emph{format\_issues}: violations of the user's format, length, count, order, language, exact-string or exclusion words, judged with the code's counts.
\item \emph{repair\_plan}: up to 12 concrete edits that change whether a requirement, need or format rule is met, each carrying the actual facts; \emph{preserve}: the parts that are correct; \emph{verdict}.
\item Rules: judge only by the user's words and the materials; never ask for caveats, disclaimers or hedges, or to weaken a committed answer; fixes must fit the user's length limits; unconstrained style is not flagged.
\end{itemize}
\paragraph{Repair (zero or one call)}
Given the checklist, the actionable part of the check with the audit's lines, the measured facts and the draft, the model revises the draft under these patch rules (verbatim):
\begin{itemize}[leftmargin=1.2em,itemsep=0pt,topsep=1pt,parsep=0pt]
\item Apply every repair\_plan item, fix every failed or partial requirement and every missing, incomplete or wrong information need with the facts and quotes in the check, and add the latent gaps; correct or remove the unsupported claims.
\item Skip a fix that the user's words or the materials contradict, and any item marked optional or ``consider''. Apply a code comparison only where the quote in the check shows the check's value; never change a draft value on evidence marked [quote not found].
\item Keep every definite answer definite: never turn an exact count, total, date or verdict into ``about'', ``approximately'', ``may'', ``appears'' or ``unverified'', and never add caveats, disclaimers or notes about sources, data quality, conflicts or what was left out, even if a fix asks for one.
\item When you change a name, number, date or verdict, change it everywhere it appears (opening sentence, table, summary) and update the totals, shares and rankings that depend on it. ``Now'' is the date the conversation states, else the latest message or record date that has already happened.
\item Patch locally: keep the draft's structure, order, headings and wording wherever the check finds no problem, and keep every part listed under preserve.
\item Make each fixed requirement visibly satisfied in the reply itself; never rely on implication.
\item Keep the user's format, length, count and language words: when the user set a limit, make room by tightening other text rather than exceeding it.
\item Keep quotations exact and every attribution; keep excluded items out of the reply.
\item No meta-commentary: never mention the check, the checklist, the draft, edits or corrections.
\end{itemize}

\footnotesize

%% file: tables/cost.tex
\footnotesize
\setlength{\tabcolsep}{2.6pt}
\begin{tabular}{@{}lrrrrrrrr@{}}
\toprule
& \multicolumn{2}{c}{Kimi K3} & \multicolumn{2}{c}{Grok 4.7} & \multicolumn{2}{c}{GLM-5} & \multicolumn{2}{c}{GPT-6-Luna} \\
\cmidrule(lr){2-3}\cmidrule(lr){4-5}\cmidrule(lr){6-7}\cmidrule(lr){8-9}
Method & Tok. & Calls & Tok. & Calls & Tok. & Calls & Tok. & Calls \\
\midrule
Vanilla & 24k & 1.0 & 28k & 1.0 & 25k & 1.0 & 24k & 1.0 \\
CCA & 72k & 4.4 & 169k & 4.7 & 63k & 3.1 & 70k & 4.0 \\
PSCI & 148k & 5.0 & 177k & 5.0 & 139k & 5.1 & 137k & 5.0 \\
Ctx2Skill & 2.54M & 163 & 3.14M & 153 & 2.37M & 158 & 3.13M$^{*}$ & 153$^{*}$ \\
\midrule
\textbf{QERA (ours)} & 128k & 3.9 & 131k & 3.8 & 112k & 4.0 & 123k & 4.0 \\
\bottomrule
\end{tabular}